\documentclass{article}

\usepackage[preprint]{neurips_2026}

\usepackage{amsthm}

\newtheorem{theorem}{Theorem}

\usepackage[utf8]{inputenc} 
\usepackage[T1]{fontenc}    
\usepackage{url}            
\usepackage{booktabs}       
\usepackage{amsfonts}       
\usepackage{nicefrac}       
\usepackage{microtype}      
\usepackage{xcolor}         
\usepackage{amsmath}
\usepackage{graphicx}
\usepackage[colorinlistoftodos]{todonotes}
\usepackage{bm}
\usepackage{multirow}
\usepackage{multicol}
\usepackage{subcaption}
\usepackage{caption}
\usepackage{algorithm}
\usepackage{algpseudocode}
\usepackage{float}
\usepackage{hyperref}
\usepackage{lineno}

\newcommand{\tht}{\bm{\theta}}

\title{Optimality of Sub-network Laplace Approximations:\\ New Results and Methods}

\author{%
  Swarnali Raha \\
  Department of Statistics\\
  University of Florida\\
  Gainesville, FL \\
  \texttt{swarnali.raha@ufl.edu} \\
  \And
   Kshitij Khare \\
   Department of Statistics \\
   University of Florida \\
   Gainesville, FL  \\
  \texttt{kdkhare@ufl.edu} \\
  \And
   Rohit K Patra \\
   Linkedin Inc. \\
   New York, NY \\
   ropatra@linkedin.com \\
}

\begin{document}

\maketitle

\begin{abstract}
  Although the Laplace approximation offers a simple route to uncertainty quantification in deep neural networks, its reliance on inverting large Hessian matrices has motivated a range of computationally feasible low-dimensional or sparse approximations. A prominent class of such methods—sub-network Laplace approximations—constructs surrogates by restricting attention to a small subset of parameters. Existing approaches in this family typically rely on diagonal, layer-wise, or other architectural heuristics for subset selection, which ignore cross-parameter interactions and lack formal optimality guarantees. In this paper, we provide a rigorous theoretical analysis of the sub-network Laplace paradigm. We prove that all sub-network Laplace methods systematically underestimate the predictive variance of the full Laplace posterior, and that this bias decreases monotonically as the retained sub-matrix expands. Leveraging this insight, we propose two principled, analytically grounded sub-network Hessian approximations: \emph{Gradient-Laplace} selects parameters with the largest average squared gradients of the model output with respect to the parameters over a reference dataset; while \emph{Greedy-Laplace} iteratively refines this selection by accounting for off-diagonal interactions in the precision matrix. We establish theoretical guarantees characterizing their optimality properties and show that Gradient-Laplace provably outperforms existing heuristic approaches. Extensive numerical studies across diverse settings indicate that these methods perform strongly relative to existing benchmarks. 
\end{abstract}

\section{Introduction} \label{sec:Introduction}

Credible measures of uncertainty for neural networks are essential for judging the reliability of their predictions and for enabling principled decision making. Frameworks that explicitly manage exploration versus exploitation---such as contextual bandits---depend critically on calibrated uncertainty estimates, and this need spans a broad set of application domains, including recommender systems~\citep{su2024long,joachims2018deep,li2010contextual}, content moderation~\citep{avadhanula2022bandits}, clinical decision support~\citep{durand2018contextual,mintz2020nonstationary,esteva2017dermatologist}, dynamic pricing~\citep{misra2019dynamic}, dialogue systems~\citep{liu2018customized}, autonomous driving~\citep{bojarski2016end}, and the identification of hallucinations in large language models~\citep{chen2023quantifying,felicioni2024importance,dwaracherla2024efficient}. Empirically, better uncertainty quantification often translates to superior downstream performance~\citep{ovadia2019can,riquelme2018deep}.

Despite this importance, scalable uncertainty quantification for deep networks remains difficult because many principled approaches are computationally demanding. A canonical example is the Laplace approximation, which places a Gaussian posterior over parameters centered at the MAP estimate $\hat{\bm{\theta}}_{\mathrm{MAP}}$ (with $\bm{\theta}$ denoting all network parameters) and uses the observed Fisher information as the precision matrix~\cite{mackay1992practical, foong2019between}. Although conceptually simple and theoretically well-motivated, practical deployment at modern scales is hindered by the need to invert a $p \times p$ matrix---with cost on the order of $O(p^3)$---when $p$ is the number of parameters.

To reduce this burden, recent work has pursued approximations to the Laplace precision matrix that preserve only its most informative structure. One direction uses low-rank surrogates, e.g., retaining only the leading eigencomponents found via Lanczos iterations~\citep{doi:10.1137/1.9780898719192}; while effective at capturing dominant curvature directions, such iterative schemes can be numerically fragile under finite precision for very large matrices~\citep{chen2024stability}. Another extreme simplifies to a diagonal precision~\citep{riquelme2018deep, zhang2020neural}, which is computationally attractive but discards all cross-parameter interactions and can therefore degrade statistical quality.

A complementary strategy concentrates the approximation on a carefully chosen subset of parameters, often referred to as a \emph{sub-network Laplace} approach~\citep{daxberger2021laplace}. Concretely, one identifies an index set $S$ of moderate size and replaces the full precision matrix $\Omega$ with a sparsified surrogate that retains only its principal submatrix $\Omega_{S,S}$, setting all remaining entries to zero. Because this surrogate is singular, its Moore--Penrose pseudoinverse is subsequently used for variance computations. The effectiveness of this approach depends critically on how the subset $S$ is selected. Existing methods largely rely on diagonal or architectural heuristics. These include restricting the approximation to final-layer parameters (\emph{NeuralLinear})~\citep{snoek2015scalable,riquelme2018deep}, which need not correspond to directions of greatest posterior variability, and selecting parameters with the smallest diagonal entries of the estimated Hessian (\emph{Subnet diagonal})~\citep[Section~5]{daxberger2021laplace}, which ignores cross-parameter correlations. Such heuristics do not account for the interaction between the gradient vector and the precision matrix $\Omega$ in the predictive variance in~\eqref{posterior:predictive:variance}. 


We develop two new subset-selection strategies that aim to retain the benefits of sub-network Laplace while addressing these limitations. The first, \emph{Gradient-Laplace}, is motivated by the form of the predictive variance (see~\eqref{posterior:predictive:variance}): parameters exhibiting larger average squared gradients over a reference dataset exert a greater influence on predictive uncertainty and are therefore prioritized in~$S$. The second, \emph{Greedy-Laplace}, builds on this strategy by additionally incorporating off-diagonal structure in the precision matrix~$\Omega$, with the objective of improving the approximation to~$\Omega^{-1}$ (see~\eqref{posterior:predictive:variance}). Greedy-Laplace begins by identifying a larger candidate subset (of twice the target size) using \emph{Gradient-Laplace}, and then refines this set via a stepwise procedure that repeatedly selects the parameter associated with the largest current diagonal entry of the evolving precision matrix, updating (i.e., shrinking) the precision through Schur complement operations to account for the parameter just chosen. 
Through extensive evaluation---including predictive approximation accuracy and  credible-interval coverage (Section~\ref{subsec:Simulation_wass} and Appendix~\ref{app:wass_real}), and Thompson-sampling regret on Wheel Bandit (Section~\ref{subsec:wheel_bandit})---we show that these methods offer tangible improvements over existing purely diagonal or structurally ordered baselines.

Also, while the {\it sub-network LA} paradigm has generated significant interest, a rigorous theoretical investigation into the quality of the resulting approximations has not been undertaken to the best of our knowledge. As a critical step in this challenging direction, in Section \ref{sec:theoretical:guarantees} we develop a rigorously justified measure of predictive variance -- called {\em Idealized Predictive Variance (IPV)} -- for any {\it sub-network LA} approach. Next, as our first theoretical contribution, we prove that {\bf any} \textit{sub-network LA} approach underestimates the (idealized) predictive variance corresponding to the full linearized Laplace based posterior, and that the discrepancy between the two decreases as more elements are added to the subset $S$ (Theorem \ref{thm1}). We also investigate high-dimensional theoretical properties of the proposed \textit{Gradient-Laplace} algorithm, and identify general settings where this algorithm uniformly outperforms other \textit{sub-network LA} algorithms (Theorem \ref{thm2}). 

The proposed methods are provided in Section \ref{sec:methods}, theoretical results are derived in Section \ref{sec:theoretical:guarantees}, and Section \ref{sec:Simulation} targets experimental validation. Proofs of the theoretical results and additional experimental details are provided in an appendix. 

\section{Setup}\label{sec:Setup}

In this work, we primarily concentrate on the \textbf{linearized Laplace approximation} (LLA)~\citep{foong2019between}. Consider a neural network that produces an output $f_{\bm{\theta}}(\mathbf{x})$ for an input $\mathbf{x}$, where the parameter vector $\bm{\theta} \in \mathbb{R}^p$ is formed by concatenating all weights and biases of the model into a single column vector. The network parameters are learned from a training set $\mathcal{D}_{\text{train}}=\{(\mathbf{x}_{n}, y_{n}) : 1 \le n \le N\}$, where $N$ denotes the total number of available training samples. We assume that the response has a Gaussian distribution centered around the target function, that is, $y_{n} \sim N(f_{\bm{\theta}}(\mathbf{x}_n),\sigma_{0}^{2})$, and we place independent Gaussian priors on the weights and biases of the network, i.e., the elements of ${\boldsymbol \theta}$.  

\subsection{Linearized Laplace Approximation} \label{subsec:LLA}
In Bayesian statistics, the Laplace Approximation is a well-known and efficient way~\citep{tierney1991laplace,mackay2003information} to approximate intractable posterior distributions. It was first used in the context of Bayesian Deep Learning in \citet{mackay1992practical}. Laplace approximation uses a Gaussian distribution, centered at the maximum a posteriori (MAP) estimate, to approximate the posterior distribution. The covariance matrix of this approximate Gaussian distribution is chosen so that the local curvature of the logarithm of the Gaussian density at the MAP estimate $\hat{\tht}_{MAP}$ matches that of the true posterior at $\hat{\tht}_{MAP}$. In particular, for the likelihood and the priors mentioned above, the Laplace Approximation approximates the posterior with $N(\hat{\tht}_{MAP},\Omega^{-1})$, where 
 
\begin{equation} \label{hessian:true}
    \Omega= - \nabla_{\tht} \nabla_{\tht} log\, p(\tht|\mathcal{D})|_{\tht=\hat{\tht}_{MAP}}. 
\end{equation}
 
 In practice, $\Omega$ is not guaranteed to be positive semidefinite, and hence, to ensure the positive semidefiniteness of the precision matrix, we replace the Hessian matrix in  \eqref{hessian:true} with the corresponding Gauss-Newton matrix. Letting ${\bf v} = (v_{1}, v_{2}, \dots, v_{p})$ denote the vector containing prior variances, and $V$ denote the diagonal matrix whose $i^{th}$ entry is given by $v_i^{-1}$, the Gauss-Newton version of the Hessian is given by:
 
\begin{align} \label{Gauss:Newton:precision}
    \Omega=\frac{1}{\sigma_{0}^{2}} \sum_{n=1}^{N}\,g(\mathbf{x}_{n})g(\mathbf{x}_{n})^{T} + V, 
\end{align}      
 
where, $g(\mathbf{x})=\nabla_{\bm{\theta}}f_{\bm{\theta}}(\mathbf{x})|_{\bm{\theta}=\hat{\bm{\theta}}_{MAP}}$, is the gradient evaluated at the MAP estimate.

Linearized Laplace approximation (LLA) was proposed by \citet{foong2019between} in this context to address the severe underfitting caused by the Laplace approximation~\citep{lawrence2001variational}. Under the setup mentioned above, the LLA provides the following approximation of the posterior predictive distribution for a new observation $(\mathbf{x}^{*},y^{*})$,
 
\begin{align} \label{posterior:predictive:variance}
  p(y^{*}|\mathbf{x}^{*},\mathcal{D}_{train}) \approx N(f_{\hat{\bm{\theta}}_{MAP}}(\mathbf{x}^{*}),\sigma_{0}^{2}+g(\mathbf{x}^{*})^{T} \Omega^{-1} g(\mathbf{x}^{*})) 
\end{align}

In the binary classification setting, we assume that $P(y_n=1|\mathbf{x}_n, \tht) = \sigma(f_{\tht}(\mathbf{x}_{n}))$. The approximate posterior distribution using Laplace approximation again centers around the MAP estimate. However, the Gauss-Newton precision matrix of the approximate distribution is now given by
 
    \begin{align}\label{Gen_Gauss:Newton:precision}
    \Omega=\sum_{n=1}^{N}\,g(\mathbf{x}_{n}) \left(- \nabla_{f}^{2}\, log \,p(\mathbf{y}_{n}|f)|_{f=f_{\hat{\bm{\theta}}_{MAP}}(\mathbf{x}_{n})}\right) g(\mathbf{x}_{n})^{T}
   + V
\end{align}
 
with $\nabla_{f}^{2}\, log \,p(\mathbf{y}_{n}|f)=-e^{f}/(1+e^{f})^{2}$.

\section{Proposed Methods} \label{sec:methods}

\noindent
Our objective is to devise principled strategies for selecting the subset $S$ used in \textit{sub-network LA}, so that the resulting approximation $[\Omega_{S,S}]^{0}$ provides a reliable surrogate for $\Omega$ in evaluating the variance term in \eqref{posterior:predictive:variance}. We outline two approaches for identifying such a subset.

\noindent
\paragraph{Gradient–Laplace Method.}The variance expression in \eqref{posterior:predictive:variance} depends simultaneously on the precision matrix $\Omega$ and the gradient vector $g(\mathbf{x}^*)$. The Gauss–Newton approximation to $\Omega$ in \eqref{Gauss:Newton:precision} is, in turn, determined by the collection $\{g(\mathbf{x}_n)\}_{n=1}^N$. In this gradient-driven selection scheme, we choose a reference dataset $\mathcal{D}$—either the training set $\mathcal{D}_{train}$ or a separate evaluation set $\mathcal{D}_{test}$—and compute the componentwise average squared gradient
 
\[
\tilde{g} = \frac{1}{|\mathcal{D}|} \sum_{\mathbf{x} \in \mathcal{D}} g^{2}(\mathbf{x}),
\]
 
where ${\bf u}^{2} := (u_{j}^{2})_{j=1}^p$ for any ${\bf u} \in \mathbb{R}^p$.  
We then form $S$ by selecting the indices corresponding to the $k$ largest entries of $\bar{g}$. We refer to this as the \emph{Gradient–Laplace method} (Algorithm~\ref{alg:neuralnet2}). This procedure requires computing an additional summary statistic of gradients. For $\mathcal{D}_{train}$ the extra cost is negligible, while for $\mathcal{D}_{test}$ it demands a backward pass on a sufficiently large subsample. The output of Algorithm~\ref{alg:neuralnet2} is the index set $S$ of cardinality $k$, and the surrogate $[\Omega_{S,S}]^{0}$ is formed by zeroing all elements of $\Omega$ outside the principal submatrix indexed by $S$. Since $[\Omega_{S,S}]^{0}$ is singular, its Moore–Penrose pseudoinverse, denoted by $\left( [\Omega_{S,S}]^{0} \right)^{-}$, is used in \eqref{posterior:predictive:variance} in place of $\Omega^{-1}$. Note that $\left( [\Omega_{S,S}]^0 \right)^-_{rs} = 
\left( \Omega_{S,S}^{-1} \right)_{rs}$ if $r,s \in S$, and all other entries of $\left( [\Omega_{S,S}]^{0} \right)^{-}$ are set to zero. 

\paragraph{Computational Complexity.}Computation of $\hat{\bm{\theta}}_{MAP}$ and the gradients is required regardless of the approximation used. Given a reference dataset $\mathcal{D}$ of size $|\mathcal{D}|$, forming $\bar{g}$ and identifying the top $k$ entries through sorting requires $O(\max\{|\mathcal{D}|p,\, p\log p\})$ operations. Constructing the chosen $k \times k$ principal submatrix $\Omega_{S,S}$ incurs an additional $O(Nk^2)$ cost (recall $N$ is the training data size). Consequently, the total computational complexity of this approach is 
$O\!\left(\max\{|\mathcal{D}|p,\, p\log p,\, N k^{2}\}\right)$. 

\begin{algorithm}[h]
\caption{Gradient-Laplace Algorithm}
\label{alg:neuralnet2}
\begin{algorithmic}[1]
\Statex \textbf{Require} $\hat{\bm{\theta}}_{MAP}$ from fitted model using 
$\mathcal{D}_{train}$; Reference data $\mathcal{D}$; user-specified
dimension $k$
\Statex {\bf Compute} gradient $g(\mathbf{x})$ for all $\mathbf{x} \in \mathcal{D}$
\Statex {\bf Compute} $\tilde{g} \in \mathbb{R}^{p}$  such that the $i$-th element is given by $\tilde{g}_{i} =\frac{1}{|\mathcal{D}|} \sum_{\mathbf{x} \in \mathcal{D}} [g_{i}(\mathbf{x})]^{2}$ 
\Statex \textbf{Define} $S$ as collection of $k$ indices corresponding to $k$ largest entries of $\tilde{g}$
\Statex \textbf{Return} $S$.
\end{algorithmic}
\end{algorithm}


    


\paragraph{Greedy–Laplace Method.}Whereas Algorithm~\ref{alg:neuralnet2} bases the selection of $S$ solely on gradient information $g$, an alternative strategy is to operate directly on the structure of the precision matrix $\Omega$. Recall that for any precision matrix, the reciprocal of its $r^{\text{th}}$ diagonal element corresponds to the conditional variance of the $r^{\text{th}}$ component given all remaining variables. Using this property, and given a desired subset size $k$, we construct a subset of $k$ parameters in a sequential, greedy fashion so as to retain those coordinates exhibiting the greatest variability.

With computational feasibility in mind (see complexity evaluation below), we first select a set of indices $S^d_{2k}$ by applying the Gradient-Laplace approach with subset size $2k$. We initialize by setting $\Omega^{(1)} = \Omega_{S^d_{2k}, S^d_{2k}}$ (with $\Omega$ defined as in \eqref{Gauss:Newton:precision} and \eqref{Gen_Gauss:Newton:precision}). At iteration $i$, we identify the coordinate associated with the largest diagonal element of $\Omega^{(i)}$ and add it to the selected subset. We then form $\Omega^{(i+1)}$ by extracting the marginal precision matrix for the remaining variables. This update requires $O((2k - i)^2)$ computations (see Step 3 of Algorithm~\ref{alg:neuralnet1}). The resulting matrix $\Omega^{(i+1)}$ has dimension $(2k-i+1) \times (2k-i+1)$. The procedure terminates after $k$ iterations. We refer to this greedy selection procedure as the \emph{Greedy–Laplace algorithm} (Algorithm~\ref{alg:neuralnet1}).
\paragraph{Computational Complexity.}In Algorithm~\ref{alg:neuralnet1}, computing and sorting the diagonal entries incurs a cost of $O(\max\{Np, p \log p\})$. Then constructing the initial estimate $\Omega^{(1)}$ requires $O(Nk^2)$ cost. Finally, Step~3 of the {\em for} loop incurs a cost of $O((2k - i + 1)^2)$ at iteration $i$. Summing over $k$ iterations, the overall computational complexity of the \emph{Greedy–Laplace} method is therefore $O(\max\{Np, p \log p, Nk^2, k^3\})$. 

\begin{algorithm}[htbp] 
\caption{Greedy-Laplace Algorithm}
\label{alg:neuralnet1}

\begin{algorithmic}
\Statex \textbf{Require} Laplace Precision matrix $\Omega$ as in (\ref{Gauss:Newton:precision}); User-specified dimension $k$
\Statex {\bf Compute} $S^d_{2k}$ = index set obtained from Gradient-Laplace approach with subset size $2k$
\Statex \textbf{Initialize} $\Omega^{(1)}=\Omega_{S^d_{2k}, S^d_{2k}}$; $S=\phi$
\Statex \For{$t=1,2,\dots,k,$}
  \Statex 
  \begin{enumerate}
      \item \textbf{Compute} $s^* = $ index of largest diagonal entry in $\Omega^{(t)}$
      \item \textbf{Update} $S=S \cup \{s^*\}$
      \item \textbf{Compute} $\Omega^{(t+1)}=\Omega^{(t)}_{-s^*,-s^*}-\frac{\Omega^{(t)}_{-s^*,s^*}\Omega^{(t)}_{s^*,-s^*}}{\Omega^{(t)}_{s^*,s^*}}$ 
  \end{enumerate}
\EndFor

\Statex \textbf{Return} $S$.
\end{algorithmic}
\end{algorithm}

\section{Some Theoretical Results and Guarantees} \label{sec:theoretical:guarantees}

\noindent
In this section, we analyze the theoretical properties of \emph{sub-network Laplace approximation (LA) methods}. We work in a regression setting with a well-specified data-generating model, where inputs $\{{\bf x}_n\}_{n=1}^N$ are drawn independently from a sub-Gaussian distribution $F_{\mathrm{in}}$ and
 
\[
y_n \sim \mathcal{N}\!\left(h_{\mathbf{true}}({\bf x}_n),\, \sigma_0^2\right),
\qquad n = 1,\dots,N.
\]

Our goal is not to establish global consistency of the MAP estimator $f_{\hat{\tht}_{\mathrm{MAP}}}(\cdot)$ or of the associated Laplace posterior as an approximation to the true regression function $h_{\mathbf{true}}(\cdot)$. Rather, we view Laplace-based methods as \emph{local} inferential tools and focus on characterizing the quality of the approximations they induce around trained solutions.

This perspective is supported by several results in relevant deep learning regimes. In the infinite-width and Neural Tangent Kernel (NTK) limits, Bayesian neural networks converge to Gaussian processes, making Laplace approximations asymptotically exact~\citep{jacot2018ntk,lee2019wide}. Related mean-field analyses show that, at large but finite width, fluctuations around deterministic training dynamics are asymptotically Gaussian, providing theoretical justification for local quadratic approximations near learned parameters~\citep{mei2018meanfield,rotskoff2018parameterization}.

Beyond these idealized regimes, there is no general theory guaranteeing global posterior validity of Laplace approximations. Overparameterization, non-identifiability, flat directions, and multimodality preclude direct application of classical Laplace theory, and Laplace posteriors should therefore be interpreted as mode-centric, local descriptions of uncertainty rather than globally accurate posterior representations. Nonetheless, in overparameterized settings, linearized Laplace approximations remain practically meaningful and theoretically motivated local approximations.

Motivated by this view, our theoretical analysis is deliberately narrow and concrete. We focus on rigorously assessing how accurately sub-network Laplace constructions approximate the full Laplace Hessian, using a principled metric that we introduce and analyze below. 

Consider a \textit{sub-network LA} method which approximates the Gauss-Newton Hessian $\Omega$ in (\ref{Gauss:Newton:precision}) by $[\Omega_{S,S}]^0$. Furthermore, recall from Section \ref{sec:Setup} that for a new observation $({\bf x}^*, y^*)$, $\sigma_{0}^{2}+g(\mathbf{x}^{*})^{T} \Omega^{-1} g(\mathbf{x}^{*})$ is the variance of the posterior predictive distribution of $y^*$. Given this, it is natural to evaluate the quality of any \textit{sub-network LA} method with output $S$, by how well the quantity 
 
\begin{equation} \label{pvdef}
    PV(S) \stackrel{\Delta}{=} g(\mathbf{x}^{*})^{T} \left( [\Omega_{S,S}]^0 \right)^{-} g(\mathbf{x}^{*})
\end{equation}
 
\noindent
approximates $PV_{full} \stackrel{\Delta}{=} g(\mathbf{x}^{*})^{T} \Omega^{-1} g(\mathbf{x}^{*})$, for a typical future input ${\bf x}^*$. Here $PV$ is an acronym for predictive variance. It follows from (\ref{Gauss:Newton:precision}) and 
(\ref{pvdef}) that 
 
\begin{equation} \label{predvarval}
PV(S) = \frac{\sigma_0^2}{N} g_S (\mathbf{x}^{*})^{T} \left( \frac{\sigma_0^2}{N} \Omega_{S,S} \right)^{-1} g_S(\mathbf{x}^{*}), 
\end{equation}

\noindent
where $g_S ({\bf x})$ denotes the subvector of $g({\bf x})$ corresponding to the indices in $S$, and 
 
\begin{equation} \label{isubmatrix}
\frac{\sigma_0^2}{N} \Omega_{S,S} = \frac{1}{N} \sum_{n=1}^N g_S({\bf x}_n) g_S({\bf x}_n)^T + \frac{\sigma_0^2}{N} V_{S,S}. 
\end{equation}

\noindent
We now construct an idealized/population version of the predictive variance metric $PV(S)$ above based on the following considerations. We assume that the maximum likelihood estimator (MLE) \( \hat{\boldsymbol \theta}_{MAP} \) converges in probability to a limit \( \bar{\boldsymbol \theta} \) as \( n \rightarrow \infty \). Importantly, this \textbf{is not} equivalent to assuming consistency—i.e., that there exists a ``true'' data-generating value of \( \tht \) to which the MLE converges. Our assumption is substantially weaker. Let $\bar{g}({\bf x}) \stackrel{\Delta}{=} \nabla_{\boldsymbol \theta} f_{\boldsymbol \theta} ({\bf x}) \mid_{{\boldsymbol \theta} = \bar{\boldsymbol \theta}}$, and let $\Lambda \stackrel{\Delta}{=} E_{{\bf x} \sim F_{in}} \left[ \bar{g}({\bf x}) \bar{g}({\bf x})^T \right]$. Replacing $g({\bf x}^*)$ in (\ref{predvarval}) by its idealized version $\bar{g}({\bf x}^*)$, and $\frac{1}{N} \sum_{n=1}^N g_S({\bf x}_n) g_S({\bf x}_n)^T$ in (\ref{isubmatrix}) by its idealized version $\Lambda_{S,S}$, we arrive at our {\bf idealized evaluation metric $IPV(\cdot)$}, defined as
 
\begin{align} \label{ipvevalcrt}
IPV(S) \stackrel{\Delta}{=} \frac{\sigma_0^2}{N} E_{{\bf x}^* \sim F_{in}} \left[ \bar{g}_S (\mathbf{x}^{*})^{T} \left(\left( \Lambda + \frac{\sigma_0^2 V}{N} \right)_{S,S}\right)^{-1} \bar{g}_S (\mathbf{x}^{*}) \right],
\end{align}    
 
and its full-Laplace based version is defined by $IPV_{full} \stackrel{\Delta}{=} IPV(\{1,2, \cdots, p\})$. 

\noindent
The precise mathematical assumptions and calculations which formally justify the use of $IPV(S)$ as an idealization of $PV(S)$ in a {\em high-dimensional setting} are presented in Appendix A.1. This metric finally leads us to a principled criterion to evaluate the accuracy of any \textit{sub-network LA} method - 

\noindent
{\bf Key evaluation criterion for any \textit{sub-network LA} method}: {\em The quality of a \textit{sub-network LA} method which results in the subset $S$ is measured by how close $IPV(S)$ (see (\ref{ipvevalcrt})) is to the full Laplace based idealized metric $IPV_{full}$. In particular, this discrepancy is measured by $\text{Dis}(S)$ defined through} 
 
\[
\text{Dis(S)} \stackrel{\Delta}{=} \left| IPV_{full} - IPV(S) \right|
\]

\noindent
Our first result establishes a somewhat intuitive - yet surprising - 
partial ordering among various $IPV(S)$ values, as $S$ varies over subsets 
of $\{1,2, \cdots, p\}$. 
\begin{theorem} \label{thm1}
Suppose $S \subseteq S' \subseteq \{1,2,\cdots,p\}$. 
Then $IPV(S) \leq IPV(S')$. In particular, $IPV(S) \leq IPV_{full}$ for 
every $S \subseteq \{1,2,\cdots,p\}$, and $\text{Dis}(S) \geq \text{Dis}(S')$ whenever $S \subseteq S'$. 
\end{theorem}

\noindent
The proof of this result is provided in Appendix \ref{proof:thm1}. The result above shows that a \textit{sub-network LA} approach always underestimates the (idealized) predictive variance of the full-Laplace approach, that $\text{Dis(S)} = IPV_{full} - IPV(S)$ (no need for absolute value in the definition), and that the discrepancy $\text{Dis}(S)$ between $IPV(S)$ and $IPV_{full}$ decreases as more elements are added to $S$. A noteworthy property of the above result is that it holds for {\bf all \textit{sub-network LA} methods}, and not just the proposed approaches in this paper. A {\em classification extension} of the IPV metric and an analogous result for Theorem \ref{thm1} in this setting are provided in Appendix \ref{ipvclass}. 

\smallskip

\noindent
We now turn to a comparison between the proposed approaches and other \textit{sub-network LA} approaches. Recall that the Gradient--Laplace approach selects the $k$ indices corresponding to the $k$ \emph{largest} entries of $\tilde{g}$ (see Algorithm~\ref{alg:neuralnet2}). In contrast, two popular existing \emph{sub-network LA} methods---\emph{Subnet Diagonal}~\citep{daxberger2021bayesian} and \emph{Last $k$}~\citep{riquelme2018deep}---choose indices in very different ways. The \emph{Subnet Diagonal} method selects the indices associated with the \emph{smallest} entries of $\tilde{g}$, while the \emph{Last $k$} method simply chooses the indices $\{1, 2, \ldots, k\}$, under the assumption that parameters are ordered from the first to the last layer of the network. 

While it is extremely difficult to provide a fully exhaustive theoretical comparison, we are able to compare the performance of the proposed {\em Gradient-Laplace} approach with other {\em sub-network LA} approaches under some instructive and meaningful constraints on $\Lambda$. 

We define $\mathcal{C}_{PI}$ as the class of matrices of the form $D + C$, where $D$ is diagonal with positive entries and $C$ is positive semi-definite and permutation-invariant. This structure for the gradient second-moment matrix is consistent with Neural Collapse results showing symmetric, isotropic structure late in training~\citep{papyan2020neural,zhu2021geometric}, arises in mean-field analyses via neuron exchangeability~\citep{mei2018meanfield}, and is supported by curvature studies identifying an isotropic bulk with a few dominant symmetric modes~\citep{martens2015kfac,ghorbani2019hessian}.

We now state our first comparison result (proof in Appendix \ref{proof:thm2}). Let $P_{true}$ denote the probability measure underlying the true data-generating model. For a given $k$, let $S_{GrL,k}$ represent the subset of size $k$ chosen by the {\em Gradient-Laplace} approach specified in 
Algorithm \ref{alg:neuralnet2}, and let $S_{SubD,k}$ represent the subset of size $k$ chosen by the {\em Subnet diagonal} approach in \cite{daxberger2021bayesian}.

\begin{theorem} \label{thm2}
Suppose $\Lambda$ belongs to the class $\mathcal{C}_{PI}$, and $V = cI_p$ for some $c > 0$ (i.e., prior variances for all elements of ${\boldsymbol \theta}$ are equal). Then, under mild concentration and separability assumptions (see Assumptions (A1), (A4), and (A6) in Appendix~\ref{app:A1}), we have
 
\begin{eqnarray*}
P_{true} \left( \text{Dis}(S_{GrL,k}) = \min_{S: |S| = k} \text{Dis}(S) \right) \to 1 \; \text{ and } \; P_{true} \left( \text{Dis}(S_{SubD,k}) = \max_{S: |S| = k} \text{Dis}(S) \right) \to 1
\end{eqnarray*}    
 
\noindent
as $N \rightarrow \infty$. In other words, the proposed {\em Gradient-Laplace} approach with subset size $k$ will {\em outperform} any other \textit{sub-network LA} approach which yields a subset of size $k$, while the {\em Subnet diagonal} approach will {\em underperform} any such approach. 
\end{theorem}

\noindent
Our second comparison result (proof in Appendix \ref{proof:thm3}) shows that the {\em Gradient-Laplace} approach outperforms a huge majority of \textit{sub-network LA} approaches (including {\em Subnet diagonal}) with high-probability for a suitable class of diagonally dominant $\Lambda$. 
\begin{theorem} \label{thm3}
Let $m_j$ denote the index of the $j^{th}$ largest diagonal entry of $\Lambda$. If $\Lambda$ is an $\epsilon$-diagonally dominant matrix (see Appendix \ref{proof:thm3}), and if $\frac{\Lambda_{m_k m_k}}{\Lambda_{m_{p-k+1} m_{p-k+1}}} > \frac{1+\epsilon}{1-\epsilon}$, then under mild concentration and separability assumptions (see Assumptions (A1), (A4), and (A6) in Appendix~\ref{app:A1}) 
 
\[
P_{true} \left( \text{Dis}(S_{GrL,k}) \leq \text{Dis}(S_{SubD,k}) \right) \to 1 \mbox{ as } N \rightarrow \infty. 
\]
 
Moreover, if $\frac{\Lambda_{m_k m_k}}{\Lambda_{m_{k+1} m_{k+1}}} > \frac{1+\epsilon}{1-\epsilon}$ then 
 
\[
P_{true} \left( \text{Dis}(S_{GrL,k}) \leq \min_{S:|S|=k, S \cap \{m_1, m_2, \cdots, m_k\} = \emptyset} \text{Dis}(S) \right) \to 1 \mbox{ as } N \rightarrow \infty. 
\]
 
\end{theorem}

\noindent
The results above highlight that when the off-diagonal entries of $\Lambda$ do not encode meaningful distinguishing structure, or are sufficiently small relative to the diagonal, the Gradient-Laplace approach is asymptotically optimal within this class. The proposed Greedy-Laplace approach is designed to improve performance when off-diagonal interactions do carry stronger additional information. This is supported by empirical evidence in Section~5. A rigorous theoretical characterization of this phenomenon remains challenging and is left for future work.

\section{Empirical Evaluation} \label{sec:Simulation}

\subsection{Experiment 1: Predictive Approximation Accuracy}
\label{subsec:predictive_approx_accuracy}
\label{subsec:Simulation_wass}

\noindent
\textbf{Experimental Setup}. We first evaluate the accuracy of each approximation in preserving the linearized Laplace posterior predictive distribution. For a test input $\mathbf{x}^{*}$, we compare the predictive distribution induced by the full precision matrix $\Omega$ with that obtained by replacing $\Omega$ with the method-specific surrogate. Since these one-dimensional Gaussian predictive distributions share the same mean, the Wasserstein-2 distance reduces to the absolute gap in predictive standard deviations,
 
\[
W_{2}(\mathbf{x}^{*})
=
\left|
\sigma_{\mathrm{Full}}(\mathbf{x}^{*})
-
\sigma_{\mathrm{method}}(\mathbf{x}^{*})
\right|.
\]

We report the average of this quantity over a fixed test subsample. This metric directly measures whether the sub-network approximation preserves the predictive uncertainty of the full Laplace posterior. We evaluate two real-data settings. 

\textbf{Setup A} is YearPredictionMSD~\citep{bertinmahieux2011yearpredictionmsd}, a tabular regression benchmark derived from the Million Song Dataset in which the goal is to predict a song's release year from $90$ audio features. Our setup follows the common Laplace-approximation benchmark protocol used in prior work~\citep{ortega2023variational}: we use the canonical train/test split, standardize the target using only the training partition, and fit a three-hidden-layer ReLU MLP with width $200$ and a scalar output head. This model has $p=98{,}801$ trainable parameters and is trained with mean squared error loss.

\textbf{Setup B} is Binary CIFAR-10, a binary image-classification benchmark constructed from CIFAR-10~\citep{krizhevsky2009learning} by collapsing the original ten classes into two balanced groups using $y_{\mathrm{binary}}=\mathbf{1}\{y\geq 5\}$. We fit CIFAR-style ResNets~\citep{he2016deep} with the standard ten-class linear head replaced by a single-logit output head, and train using binary cross-entropy loss. The headline result uses ResNet-110 with $p=1{,}730{,}129$ parameters; smaller ResNet-20/32/56 backbones are reported in Appendix~\ref{app:wass_real:scaling}.


For Setup B, the single-logit Bernoulli likelihood reduces the Gauss--Newton precision in~\eqref{Gen_Gauss:Newton:precision} to the same outer-product form as in regression, but with each training Jacobian weighted by the scalar factor $p_n(1-p_n)$, where $p_n=\sigma(f_{\theta}(x_n))$. Thus the sub-network Laplace methods apply without modification after replacing the ordinary Jacobian by the Hessian-weighted Jacobian. 

We compare \emph{Gradient-Laplace} and \emph{Greedy-Laplace} against \emph{Subnet Diagonal}~\citep{daxberger2021bayesian}, \emph{Last $k$}, and  \emph{NeuralLinear} \citep{riquelme2018deep}. Definitions of the  benchmark algorithms are provided in Appendix~\ref{app:benchmark_algo}. Within each seed, all sub-network methods share the same trained MAP estimate; only the selected index set $S$ differs. Results are averaged over ten independent random seeds, with standard errors reported across seeds. For both setups, full architecture, training, and hyperparameter details are provided in Appendix~\ref{app:wass_real}.  

\begin{figure}[h]
    \centering
    \includegraphics[width=.9\linewidth]{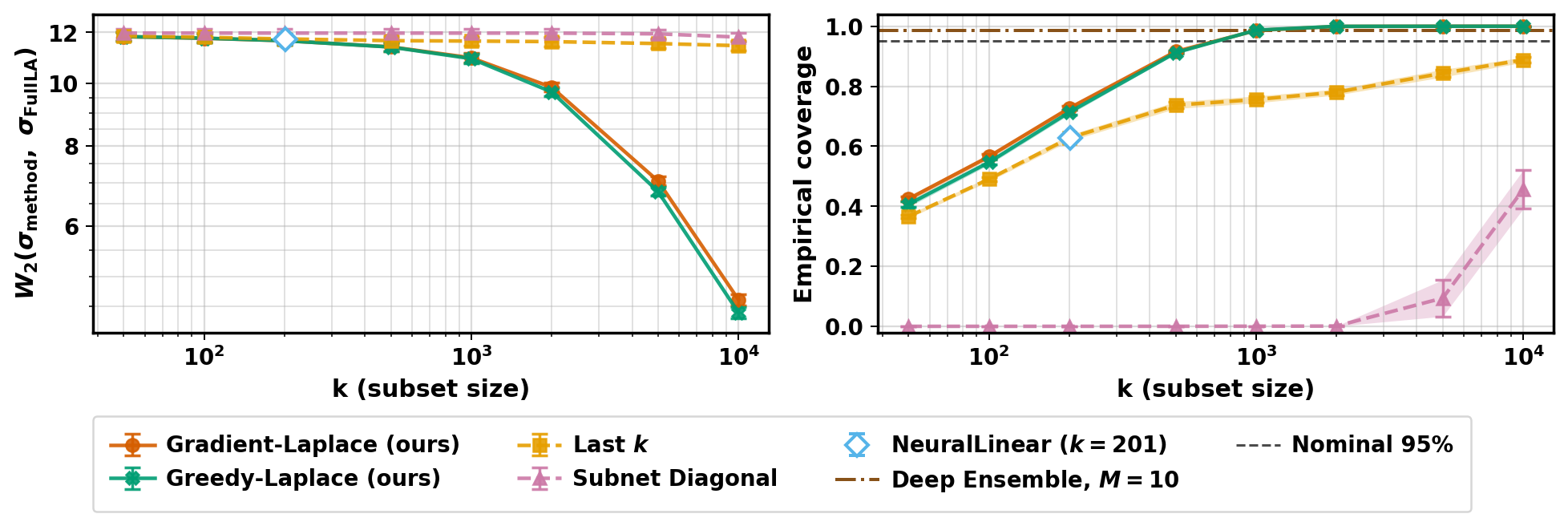}
    \caption{Setup A (\textbf{YEAR Prediction MSD}, regression, $p=98{,}801$). \textbf{Left:} per-test-point Wasserstein distance between full Laplace predictive and its sub-network surrogate as a function of subset size $k$; lower is better. \textbf{Right:} calibration diagnostic: empirical coverage of nominal $95\%$ posterior credible intervals. Lines show means over ten seeds and shaded bands show one standard error.}
    \label{fig:wass_year}
\end{figure}

\begin{figure}[h]
    \centering
    \includegraphics[width=.9\linewidth]{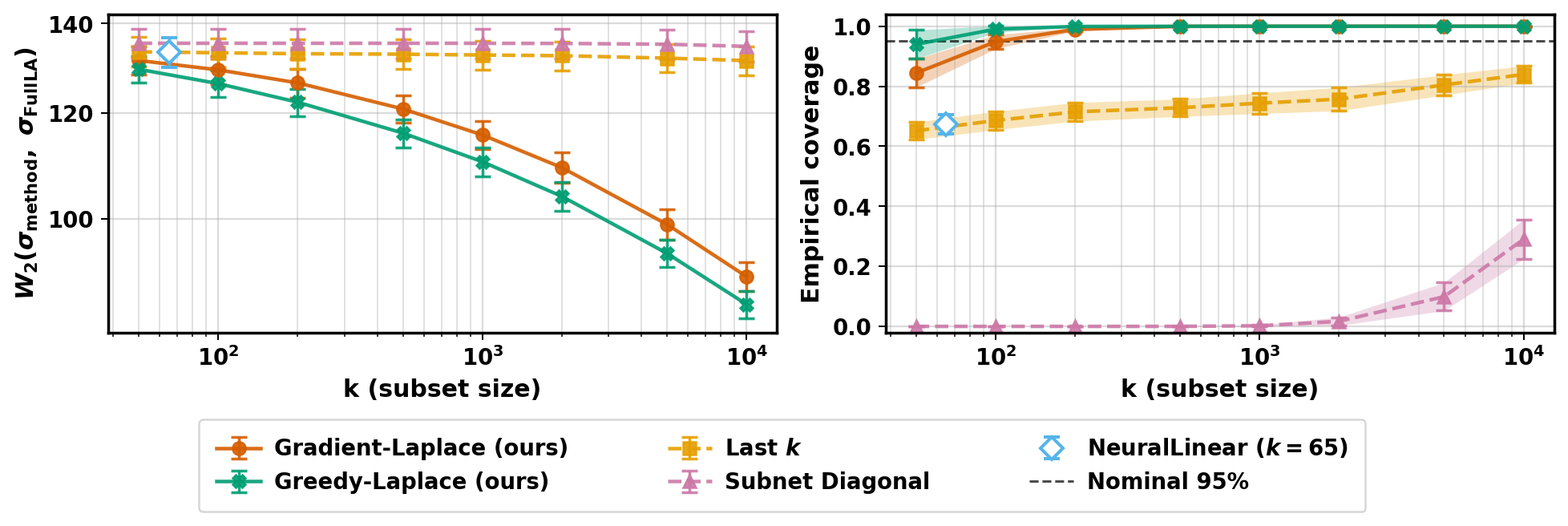}
    \caption{Setup B (\textbf{Binary CIFAR-10}, ResNet-110, $p=1{,}730{,}129$). \textbf{Left:} per-test-point average Wasserstein distance between the full Laplace predictive and its sub-network surrogate as a function of subset size $k$; lower is better. \textbf{Right:} secondary calibration diagnostic: empirical coverage of nominal $95\%$ posterior credible intervals. Lines show means over ten independent random seeds and shaded bands show one standard error. The Smaller backbones reported in Appendix~\ref{app:wass_real:scaling}.}
    \label{fig:wass_cifar_binary}
\end{figure}

\noindent
\textbf{Results and Discussion}. Figures~\ref{fig:wass_year} and~\ref{fig:wass_cifar_binary} show that the proposed methods provide substantially closer posterior predictive approximations than existing sub-network baselines. On YEAR, both \emph{Gradient-Laplace} and \emph{Greedy-Laplace} sharply reduce the predictive-standard-deviation gap as $k$ increases, while \emph{Subnet Diagonal} and \emph{Last $k$} remain much flatter. \emph{NeuralLinear}, which corresponds to a fixed last-layer subset ($k=201$ for YEAR and $k=65$ for the binary CIFAR-10 head), is matched or surpassed by the proposed methods at comparable subset sizes.

On Binary CIFAR-10 with ResNet-110, \emph{Greedy-Laplace} consistently improves over \emph{Gradient-Laplace}, with the largest gains at moderate and large $k$. This advantage becomes more visible as the backbone grows (see Appendix~\ref{app:wass_real:scaling}): the two methods are nearly tied on ResNet-20, a small gap opens on ResNet-32/56, and the gap is clearest on ResNet-110. This pattern is consistent with the role of \emph{Greedy-Laplace}: after the gradient-based screening step, its Schur-complement refinement accounts for off-diagonal precision structure that \emph{Gradient-Laplace} ignores.

The right panels of Figures~\ref{fig:wass_year} and~\ref{fig:wass_cifar_binary} provide a secondary coverage check using the same predictive variances as in the Wasserstein comparison. On YEAR, the proposed methods approach the \emph{Deep Ensemble} reference at sufficiently large $k$, while \emph{Subnet Diagonal} under-covers across all subset sizes. On Binary CIFAR-10, a similar pattern holds: \emph{Gradient-Laplace} and \emph{Greedy-Laplace} provide more reliable coverage than existing sub-network baselines. These results are presented as a calibration check for the predictive-variance approximation rather than a separate experiment.

Additional experiments in Appendix~\ref{app:wass_real} further stress-test this conclusion through a CIFAR-style ResNet backbone sweep, a full ten-class CIFAR-10 experiment using the proper multi-class softmax-Hessian formulation, and a four-dataset UCI regression suite. Across these settings, the conclusion remains stable: \emph{Gradient-Laplace} and \emph{Greedy-Laplace} consistently provide closer approximations to the full Laplace predictive distribution than diagonal or architecture-based baselines. Overall, these results support the central empirical claim that selecting parameters by their contribution to predictive uncertainty yields more faithful sub-network Laplace approximations, consistent with Theorems~\ref{thm2} and~\ref{thm3}  in Section~\ref{sec:theoretical:guarantees}. 



\subsection{Experiment 2: Wheel Bandit}
\label{subsec:wheel_bandit}

We next evaluate whether the proposed sub-network Laplace approximations support effective exploration in a sequential decision-making problem. We use the \emph{Wheel Bandit} benchmark of~\citet{riquelme2018deep}, a sparse-reward contextual bandit used to test Bayesian neural-network approximations under Thompson sampling. Contexts are sampled from the unit disk, with high rewards occurring only for the correct outer arm when the context lies in a thin annulus near the boundary. We set $\delta = 0.95$, so the high-reward region occupies only $1-\delta^{2} = 9.75\%$ of the disk. This regime is challenging, as low regret requires sufficiently informative uncertainty estimates to discover rare high-reward regions. Full details and additional experiment with and $\delta =.9$ are in Appendix~\ref{app:wheel_bandit} .

All methods are evaluated within the Neural Thompson Sampling framework~\citep{zhang2020neural}. At each round, the agent samples a reward for each arm from the posterior predictive distribution and selects the arm with the largest sampled value. For sub-network LA methods, the predictive variance is computed from the regression-form linearized Laplace approximation with the precision restricted to the selected subset. We compare \emph{Gradient-Laplace} and \emph{Greedy-Laplace} against \emph{Subnet Diagonal}, \emph{Last $k$}, \emph{NeuralLinear}, and a deterministic MAP baseline (details provided in Appendix~\ref{app:benchmark_algo}). 

The reward model is a fully connected neural network with two hidden layers of $100$ ReLU units each and a scalar reward prediction head. The input is the two-dimensional context concatenated with a five-dimensional one-hot encoding of the arm, giving input dimension $7$ and total parameter count $p = 11{,}001$. We use a horizon of $T=16{,}000$ rounds and report final cumulative regret averaged over 25 random seeds. For sub-network methods, we evaluate $k \in \{100,500\}$. Training and hyperparameter details are deferred to Appendix~\ref{app:wheel_bandit}.

\begin{figure}[h]
    \centering
    \includegraphics[width=0.78\linewidth]{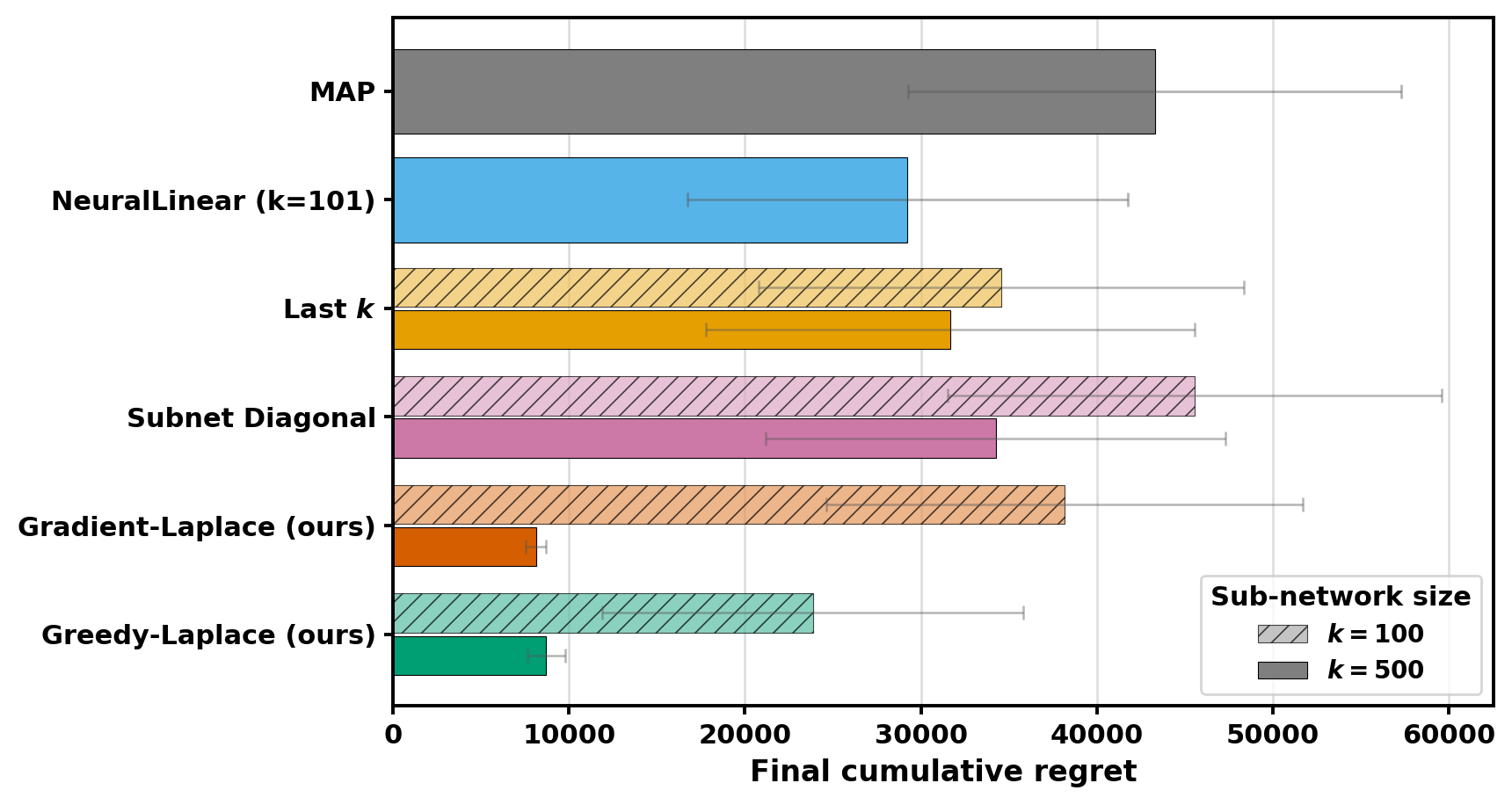}
    \caption{Final cumulative regret on the Wheel Bandit at $\delta=0.95$ over $T=16{,}000$ rounds.  Lower is better.  Bars show means over 25 seeds with $\pm 1.96\,\mathrm{SE}$ error bars. Paired bars denote $k=100$ and $k=500$ for sub-network methods.}
    \label{fig:Creg_plot}
\end{figure}

Figure~\ref{fig:Creg_plot} shows that the proposed methods substantially reduce regret relative to existing sub-network baselines. At $k=100$, \emph{Greedy-Laplace} achieves the lowest regret among sub-network methods, improving over both \emph{Subnet Diagonal} and \emph{Last $k$}. At $k=500$, \emph{Gradient-Laplace} and \emph{Greedy-Laplace} are statistically comparable and both sharply outperform the existing sub-network baselines. The deterministic MAP baseline performs poorly, confirming that posterior-driven exploration is essential in this sparse-reward regime. Overall, these results show that preserving more informative Laplace posterior directions translates into more significantly improved learning.

\noindent
{\bf Acknowledgement}. Khare's work on this paper was supported by NSF-DMS-2410677. 

\bibliographystyle{plainnat}
\bibliography{references}


\newpage

\appendix

\section{Proofs of the Theoretical Results} \label{app:thm_proofs}

\noindent
In this section, we present rigorous arguments to support the heuristics that lead to the definition of the idealized predictive variance metric 
 
\[
IPV(S) = \frac{\sigma_{0}^{2}}{N}E_{{\bf x}^* \sim F_{in}} \left[ \bar{g}_S (\mathbf{x}^{*})^{T} \left( \Lambda_{S,S} + \frac{\sigma_{0}^{2} V_{S,S}}{N} \right)^{-1} \bar{g}_S (\mathbf{x}^{*}) \right],      
\]

\noindent
and also provide the proofs of Theorems 1 and 2. 

\subsection{Justification for the idealized metric $IPV$} \label{app:A1}

\noindent
Next, we present the precise assumptions and justification for using the idealized predictive variance metric 
 
\[
IPV(S) = \frac{\sigma_0^2}{N} E_{{\bf x}^* \sim F_{in}} \left[ \bar{g}_S (\mathbf{x}^{*})^{T} \left( \Lambda_{S,S} + \frac{\sigma_0^2 V_{S,S}}{N} \right)^{-1} \bar{g}_S (\mathbf{x}^{*}) \right]
\]

\noindent
as an approximation for 
 
\[
E_{{\bf x}^* \sim F_{in}} [PV(S)] = \frac{\sigma_0^2}{N} E_{{\bf x}^* \sim F_{in}} \left[ g_S(\mathbf{x}^{*})^{T} \left( \frac{1}{N} \sum_{n=1}^N g_S({\bf x}_n) g_S({\bf x}_n)^T + \frac{\sigma_0^2 V_{S,S}}{N} \right)^{-1} g_S(\mathbf{x}^{*}) \right]. 
\]

\noindent
For convenience, we assume $\sigma_0^2 = 1$ without loss of generality. We will also assume that $V = I_p$, this can be relaxed to assuming that the diagonal entries $v_i$ are uniformly bounded. {\bf We work in a growing dimensions setting}, i.e., we allow for $p$, the total number of weights and biases, and $|S| = k$, the sub-network size, to grow with the sample size $N$. We assume the following. 
\begin{itemize}
    \item (A1) $\sqrt{\frac{k^3 \log p}{N}} \rightarrow 0$ as $N \rightarrow \infty$. 
    \item (A2) There exists $\bar{\boldsymbol \theta}$ such that $\alpha_n := k \|\hat{\boldsymbol \theta}_{MAP} - \bar{\boldsymbol \theta}\|_{max} = o_{P_{true}} (1/k)$, where $P_{true}$ denotes the probability measure underlying the true data generating model, and $\|\cdot\|_{max}$ denotes the maximum absolute value norm.  
    \item (A3) Let 
     
        \[
        H({\boldsymbol \theta}, {\bf x}) \stackrel{\Delta}{=} \nabla_{\bm{\theta}} \nabla_{\bm{\theta}}f_{\bm{\theta}}(\mathbf{x}).  
        \]

    \noindent
    We assume that $\sup_{{\boldsymbol \theta} \in B_\epsilon(\bar{\boldsymbol \theta})} \|H({\boldsymbol \theta}, {\bf x})\|_{\infty}$ has a uniformly bounded second moment (in $N$, when ${\bf x} \sim F_{in}$) for some fixed $\epsilon \in (0,1)$. Here $B_\epsilon(\bar{\boldsymbol \theta})$ represents a ball of radius $\epsilon$ around $\bar{\boldsymbol \theta}$, and $\|\cdot\|_{\infty}$ is the maximum absolute row sum norm (see \cite{katsevich2025unified} for similar assumptions in the context of high-dimensional Laplace approximations).
    \item (A4) We assume that $\bar{g}({\bf x})$ (when $x \sim F_{in}$) is sub-Gaussian with $\Lambda^{-1/2} \bar{g}({\bf x})$ having a uniformly bounded sub-Gaussian norm (in $N$). This can be achieved, for example, by assuming that $F_{in}$ is a Gaussian distribution, and $g$ is a Lipschitz function with uniformly bounded Lipschitz norm (see Equation (2.4) in \cite{Ledoux:1996}). 
   \item (A5) We assume that the eigenvalues of $\Lambda$ are uniformly bounded above in $N$. 
     
\end{itemize}

\noindent
Now, using the sub-Gaussianity of $\bar{g}({\bf x})$ along with Assumptions (A4), and the Hanson-Wright inequality~\citep{Rudelson:Vershynin:2013} and Lemma F.2 in \citet{10.1214/15-AOS1315}, it follows that 
 
\begin{equation} \label{HSbound}
\sup_{S: \; |S| = k} \frac{\left\| \frac{1}{N} \sum_{n=1}^N \bar{g}_S({\bf x}_n) \bar{g}_S({\bf x}_n)^T  - \Lambda_{S,S} \right\|}{\|\Lambda_{S,S}\|_2} = O_{P_{true}} \left( \sqrt{\frac{k \log p}{N}} \right). 
\end{equation}

\noindent
Next, using the mean value theorem for the function $\nabla_{\boldsymbol \theta} f_{\boldsymbol \theta} ({\bf x})$ (as a function of $\tht$) and 
Assumption (A2) (which implies that $\hat{\boldsymbol \theta}_{MAP} \in B_\epsilon(\bar{\boldsymbol \theta})$ with high probability for large enough $N$), it follows that on a set with $P_{true}$-probability tending to one, we have 
 
\begin{equation} \label{maxbound}
\|g({\bf x}) - \bar{g}({\bf x})\|_{max} \leq \|\hat{\boldsymbol \theta}_{MAP} - \bar{\boldsymbol \theta}\|_{max} \sup_{{\boldsymbol \theta} \in B_\epsilon(\bar{\boldsymbol \theta})} \|H({\boldsymbol \theta}, {\bf x}) \|_\infty.  
\end{equation}

\noindent
for every $\mathbf{x}$. Now fix any $S$ with $|S| = k$. Since 
 
\begin{eqnarray*}
& & \left\| g_S({\bf x}_n) g_S({\bf x}_n)^T - \bar{g}_S({\bf x}_n) \bar{g}_S({\bf x}_n)^T \right\|\\
&\leq& \left\| \left( g_S({\bf x}_n) - \bar{g}_S({\bf x}_n) \right) \bar{g}_S({\bf x}_n)^T \right\| + \left\| {g}_S({\bf x}_n) \left( g_S({\bf x}_n) - \bar{g}_S({\bf x}_n) \right)^T \right\|\\
&\leq& \left\| \left( g_S({\bf x}_n) - \bar{g}_S({\bf x}_n) \right) \bar{g}_S({\bf x}_n)^T \right\| + \left\|\bar{g}_S({\bf x}_n) \left( g_S({\bf x}_n) - \bar{g}_S({\bf x}_n) \right)^T \right\| +\\
& & \left\|\left( {g}_S({\bf x}_n) - \bar{g}_S({\bf x}_n) \right) \left( g_S({\bf x}_n) - \bar{g}_S({\bf x}_n) \right)^T \right\|, 
\end{eqnarray*}

\noindent
it follows by Assumption (A2) that with $P_{true}$-probability tending to one, 
 
\begin{eqnarray}
& & \left\| \frac{1}{N} \sum_{n=1}^N g_S({\bf x}_n) g_S({\bf x}_n)^T - \frac{1}{N} \sum_{n=1}^N \bar{g}_S({\bf x}_n) \bar{g}_S({\bf x}_n)^T \right\|\\
&\leq& 2 \sqrt{k} \|\hat{\boldsymbol \theta}_{MAP} - \bar{\boldsymbol \theta}\|_{max} \frac{1}{N} \sum_{n=1}^N \|\bar{g}_S ({\bf x}_n)\| \sup_{{\boldsymbol \theta} \in B_\epsilon(\bar{\boldsymbol \theta})} \|H({\boldsymbol \theta}, {\bf x}_n) \|_\infty + \nonumber\\
& & k \|\hat{\boldsymbol \theta}_{MAP} - \bar{\boldsymbol \theta}\|_{max}^2 \frac{1}{N} \sum_{n=1}^N \left( \sup_{{\boldsymbol \theta} \in B_\epsilon(\bar{\boldsymbol \theta})} \|H({\boldsymbol \theta}, {\bf x}_n) \|_\infty \right)^2\\
&\leq& 2 \sqrt{k} \|\hat{\boldsymbol \theta}_{MAP} - \bar{\boldsymbol \theta}\|_{max} \times \nonumber\\
& & \sqrt{\frac{1}{N} \sum_{n=1}^N \|\bar{g}_S ({\bf x}_n)\|^2} \sqrt{\frac{1}{N} \sum_{n=1}^N \left( \sup_{{\boldsymbol \theta} \in B_\epsilon(\bar{\boldsymbol \theta})} \|H({\boldsymbol \theta}, {\bf x}_n) \|_\infty \right)^2} + \nonumber\\
& & k \|\hat{\boldsymbol \theta}_{MAP} - \bar{\boldsymbol \theta}\|_{max}^2 \frac{1}{N} \sum_{n=1}^N \left( \sup_{{\boldsymbol \theta} \in B_\epsilon(\bar{\boldsymbol \theta})} \|H({\boldsymbol \theta}, {\bf x}_n) \|_\infty \right)^2\\
&=& \sqrt{\|\Lambda_{S,S}\|} O_{P_{true}} \left( \alpha_n \right) + O_{P_{true}} \left( \frac{\alpha_n^2}{k} \right)
\label{limitbound}
\end{eqnarray}

\noindent
The last step follows by Assumption (A3) and a straightforward appllication of Chebyshev's inequality. Combining (\ref{HSbound}) and (\ref{limitbound}), and using Assumptions (A1) and (A2), we get
 
\begin{eqnarray} \label{matcov}
& & \left\| \frac{1}{N} \sum_{n=1}^N {g}_S({\bf x}_n) {g}_S({\bf x}_n)^T  - \Lambda_{S,S} \right\| \nonumber\\
&=& \sqrt{\|\Lambda_{S,S}\|} O_{P_{true}} \left( \alpha_n \right) + \|\Lambda_{S,S}\| O_{P_{true}} \left( \sqrt{\frac{k \log p}{N}} \right) + O_{P_{true}} \left( \frac{\alpha_n^2}{k} \right). 
\end{eqnarray}

\noindent
By Assumptions (A1) and (A5), the identity $\|A^{-1} - B^{-1}\| \leq \|A^{-1}\|\|A - B\|\|B^{-1}\|$, along with 
 
\[
\left\| \left( \frac{1}{N} \sum_{n=1}^N {g}_S({\bf x}_n) {g}_S({\bf x}_n)^T + \frac{ V_{S,S}}{N} \right)^{-1} \right\| \leq N, 
\]

\noindent
and 
 
\[
\left\| \left( \Lambda_{S,S} + \frac{ V_{S,S}}{N} \right)^{-1} \right\| \leq \left\| \Lambda_{S,S}^{-1} \right\|, 
\]

\noindent
it follows that 
 
\begin{eqnarray} \label{doublebd}
& & \left\| \left( \frac{1}{N} \sum_{n=1}^N {g}_S({\bf x}_n) {g}_S({\bf x}_n)^T + \frac{ V_{S,S}}{N} \right)^{-1} - \left( \Lambda_{S,S} + \frac{ V_{S,S}}{N} \right)^{-1} \right\| \nonumber\\
&=& N \|\Lambda_{S,S}^{-1}\| \left( \sqrt{\|\Lambda_{S,S}\|} O_{P_{true}} \left( \alpha_n \right) + \|\Lambda_{S,S}\| O_{P_{true}} \left( \sqrt{\frac{k \log p}{N}} \right) + O_{P_{true}} \left( \frac{\alpha_n^2}{k} \right) \right)
\end{eqnarray}

\noindent
Note that 
 
\begin{align}\label{qfbound}
&  |{\bf r}^T A^{-1} {\bf r} - {\bf s}^T B^{-1} {\bf s}| \nonumber\\
&\leq  (\|{\bf r}\| + \|{\bf s}\|)\|A^{-1}\|\|{\bf r} - {\bf s}\| + \|{\bf s}\|^2 \|A^{-1} - B^{-1}\| \nonumber\\
&\leq (\|{\bf r}\| + \|{\bf s}\|)\|A^{-1} - B^{-1}\|\|{\bf r} - {\bf s}\| + (\|{\bf r}\| + \|{\bf s}\|)\|B^{-1}\|\|{\bf r} - {\bf s}\| +  \notag \\
& \quad  + \|{\bf s}\|^2 \|A^{-1} - B^{-1}\|
\end{align}

\noindent
with ${\bf r} = g_S ({\bf x}^*)$, ${\bf s} = \bar{g}_S ({\bf x}^*)$, $A = \frac{1}{N} \sum_{n=1}^N {g}_S({\bf x}_n) {g}_S({\bf x}_n)^T + \frac{ V_{S,S}}{N}$ and $B = \Lambda_{S,S} + \frac{ V_{S,S}}{N}$. By (\ref{maxbound}), we get
 
\begin{eqnarray*}
E_{{\bf x}^* \sim F_{in}} \left[ \|g_S ({\bf x}^*) - \bar{g}_S ({\bf x}^*)\|\right] 
&\leq& \sqrt{k} E_{{\bf x}^* \sim F_{in}} \left[ \|\hat{\boldsymbol \theta}_{MAP} - \bar{\boldsymbol \theta}\|_{max} \sup_{{\boldsymbol \theta} \in B_\epsilon(\bar{\boldsymbol \theta})} \|H({\boldsymbol \theta}, {\bf x}^*) \|_\infty \right]. 
\end{eqnarray*}

\noindent
Using Assumptions (A2) and (A3), we obtain 
 
\[
E_{{\bf x}^* \sim F_{in}} \left[ \|g_S ({\bf x}^*) - \bar{g}_S ({\bf x}^*)\|\right] = O_{P_{true}} \left( \frac{\alpha_n}{\sqrt{k}} \right)
\]

\noindent
Using this along with Assumption (A5), it follows that
 
\[
E_{{\bf x}^* \sim F_{in}} \left[ \|g_S ({\bf x}^*)\| + \|\bar{g}_S ({\bf x}^*)\|\right] = O_{P_{true}} (\sqrt{k}). 
\]

\noindent
Combining these observations with (\ref{doublebd}) and (\ref{qfbound}), and recalling the additional factor of $\frac{\sigma_0^2}{N}$ factor in both $PV(S)$ and $IPV(S)$, we obtain 

 
\begin{eqnarray*}
& & \left| E_{{\bf x}^* \sim F_{in}} \left[ PV(S) \right] - IPV(S) \right| \nonumber\\ 
&=& \|\Lambda_{S,S}^{-1}\| \left( \sqrt{\|\Lambda_{S,S}\|} O_{P_{true}} \left( k \alpha_n \right) + \|\Lambda_{S,S}\| O_{P_{true}} \left( \sqrt{\frac{k^3 \log p}{N}} \right) + O_{P_{true}} \left( \alpha_n^2 + \frac{\alpha_n}{N} \right) \right). 
\end{eqnarray*}

\noindent
 Furthermore, the $o_{P_{true}}$ and $O_{P_{true}}$ statements hold uniformly over all $S$ with $|S| = k$. Hence, as long as the RHS vanishes in the limit, the idealization of $PV(S)$ with $IPV(S)$ is rigorously well-justified. One sufficient (but clearly not necessary) condition for this would be the existence of a uniform positive lower bound for the eigenvalues of $\Lambda$. 

\subsection{Proof of Theorem 1} \label{proof:thm1}

\noindent
First note that for any subset $S$ of $\{1,2, \cdots, p\}$, we have
\begin{linenomath*}
\begin{eqnarray}
IPV(S) 
&=& \frac{\sigma_0^2}{N} E_{{\bf x}^* \sim F_{in}} \left[ \bar{g}_S (\mathbf{x}^{*})^{T} \left( \Lambda_{S,S} + \frac{\sigma_0^2 V_{S,S}}{N} \right)^{-1} \bar{g}_S (\mathbf{x}^{*}) \right] \nonumber\\
&=& \frac{\sigma_0^2}{N} E_{{\bf x}^* \sim F_{in}} \left[ tr \left( \bar{g}_S (\mathbf{x}^{*})^{T} \left( \Lambda_{S,S} + \frac{\sigma_{0}^{2} V_{S,S}}{N} \right)^{-1} \bar{g}_S (\mathbf{x}^{*}) \right) \right] \nonumber\\
&=& \frac{\sigma_0^2}{N} E_{{\bf x}^* \sim F_{in}} \left[ tr \left( \left( \Lambda_{S,S} + \frac{\sigma_0^2 V_{S,S}}{N} \right)^{-1} \bar{g}_S (\mathbf{x}^{*}) \bar{g}_S (\mathbf{x}^{*})^{T} \right) \right] \nonumber\\
&=& \frac{\sigma_0^2}{N} \,tr \left( \left( \Lambda_{S,S} + \frac{\sigma_0^2 V_{S,S}}{N} \right)^{-1} E_{{\bf x}^* \sim F_{in}} \left[ \bar{g}_S (\mathbf{x}^{*}) \bar{g}_S (\mathbf{x}^{*})^{T} \right] \right) \nonumber\\
&=& \frac{\sigma_0^2}{N} \,tr \left( \left( \Lambda_{S,S} + \frac{\sigma_0^2 V_{S,S}}{N} \right)^{-1} \Lambda_{S,S} \right). \label{tridentity}
\end{eqnarray}    
\end{linenomath*}

\noindent
Now consider two subsets $S, S'$ such  that $S \subseteq S'$. Let $\widetilde{S} = S' \setminus S$. 
Without loss of generality, let 
\begin{linenomath*}
\[
\Lambda_{S',S'} = \left[ \begin{matrix}
    A & B \cr
    B^T & C
\end{matrix} \right] \hspace{0.2in} and \hspace{0.2in} \frac{\sigma_0^2 V_{S',S'}}{N} = \left[ \begin{matrix}
    D & {\bf 0} \cr
    {\bf 0} & E 
\end{matrix}\right], 
\]
\end{linenomath*}

\noindent
where $A = \Lambda_{S,S}$, $B = \Lambda_{S,\widetilde{S}}$, $C = \Lambda_{\widetilde{S}, \widetilde{S}}$, $D = \frac{\sigma_0^2 V_{S,S}}{N}$ and $E = \frac{\sigma_0^2 V_{\widetilde{S},\widetilde{S}}}{N}$. We use ${\bf 0}$ to denote any matrix of all zero entries, the dimension of the matrix will be obvious from the context throughout our presentation. It follows that 
\begin{linenomath*}
\begin{eqnarray*}
    \frac{N}{\sigma_0^2} IPV(S') 
    &=& tr \left( \left( \Lambda_{S',S'} + \frac{\sigma_0^2 V_{S',S'}}{N} \right)^{-1} \Lambda_{S',S'} \right)\\
    &=& tr \left( \left( \left[ \begin{matrix}
    A & B \cr
    B^T & C
\end{matrix} \right] + \left[ \begin{matrix}
    D & {\bf 0} \cr
    {\bf 0} & E 
\end{matrix}\right] \right)^{-1} \left[ \begin{matrix}
    A & B \cr
    B^T & C
\end{matrix} \right] \right). 
\end{eqnarray*}    
\end{linenomath*}

\noindent
Using the form of the inverse of a partitioned matrix, using $\tilde{A} = A + D$, $\tilde{C} = C + E$ and $F = \tilde{C} - B^T \tilde{A}^{-1} B$, it follows that 
\begin{linenomath*}
\begin{eqnarray*}
    \frac{N}{\sigma_0^2} IPV(S') 
    &=& tr \left( \left[ \begin{matrix}
    \tilde{A}^{-1} + \tilde{A}^{-1} B F^{-1} B^T \tilde{A}^{-1} & -\tilde{A}^{-1} BF^{-1} \cr
    -F^{-1} B^T \tilde{A}^{-1} & F^{-1}
\end{matrix} \right] \left[ \begin{matrix}
    A & B \cr
    B^T & C
\end{matrix} \right] \right)\\
    &=& tr \left( \tilde{A}^{-1} A + \tilde{A}^{-1} B F^{-1} B^T \tilde{A}^{-1} A - \tilde{A}^{-1} B F^{-1} B^T \right) + tr \left( F^{-1} C - F^{-1} B^T \tilde{A}^{-1} B \right)\\
    &=& tr \left( \tilde{A}^{-1} A \right) + tr \left( F^{-1} B^T \tilde{A}^{-1} A \tilde{A}^{-1} B \right) - 2 tr \left( F^{-1} B^T \tilde{A}^{-1} B \right) + tr \left( F^{-1} C \right)\\
    &=& \frac{N}{\sigma_0^2} IPV(S) + tr \left( F^{-1} B^T \left( \tilde{A}^{-1} A \tilde{A}^{-1} - 2 \tilde{A}^{-1} + A^{-1} \right) B \right) + tr \left( F^{-1} \left( C - B^T A^{-1} B \right) \right). 
\end{eqnarray*}    
\end{linenomath*}

\noindent
To establish the required result, it suffices to show that 
$C - 2 B^T \tilde{A}^{-1} B + B^T \tilde{A}^{-1} A \tilde{A}^{-1} B$ is positive semi-definite. We will show that by expressing this matrix as a limit for positive semi-definite matrices. For every $n \geq 1$, let $\tilde{A}_n := A + \frac{1}{n} D$. Then, with 
$\Gamma_n := \tilde{A}_n^{1/2} \tilde{A}^{-1} \tilde{A}_n^{1/2}$, we have 
\begin{eqnarray*}
 & & C - 2 B^T \tilde{A}^{-1} B + B^T \tilde{A}^{-1} \tilde{A}_n \tilde{A}^{-1} B\\
 &=& \left( C - B^T \tilde{A}_n^{-1} B \right) + B^T \left( \tilde{A}^{-1} \tilde{A}_n \tilde{A}^{-1} - 2 \tilde{A}^{-1} + \tilde{A}_n^{-1} \right) B\\
 &=& \left( C - B^T \tilde{A}_n^{-1} B \right) + B^T \tilde{A}_n^{-1/2} \left( \Gamma_n^2 - 2 \Gamma_n + I \right) \tilde{A}_n^{-1/2} B\\
 &=& \left( C - B^T \tilde{A}_n^{-1} B \right) + B^T \tilde{A}_n^{-1/2} \left( \Gamma_n - I \right) \left( \Gamma_n - I \right) \tilde{A}_n^{-1/2} B. 
\end{eqnarray*}

\noindent
Since $\tilde{A}_n$ and $\Gamma_n$ are symmetric matrices, it follows that $C - 2 B^T \tilde{A}^{-1} B + B^T \tilde{A}^{-1} \tilde{A}_n \tilde{A}^{-1} B$ is positive semi-definite. Taking the limit $n \rightarrow \infty$, we obtain $C - 2 B^T \tilde{A}^{-1} B + B^T \tilde{A}^{-1} A \tilde{A}^{-1} B$ is positive semi-definite. Hence $IPV(S) \leq IPV(S')$. \hfill$\Box$

\subsection{Classification Extension of IPV and Theorem 1} \label{ipvclass}

In regression, $PV(S)$ measures epistemic predictive variance, and normalizing by aleatoric uncertainty (which is the constant $\sigma_0^2$) does not affect comparisons. In classification, however, aleatoric uncertainty depends on the input, making such normalization essential. Similar normalizations appear in BALD \citep{houlsby2011bald}, as well as in Fisher-normalized uncertainty measures dating back to \citet{mackay1992practical}; see also \citet{kendall2017uncertainties}. 

For binary classification, with a future observation $\tilde{y} \mid \boldsymbol{\theta}, \tilde{\mathbf{x}} \sim \mathrm{Bernoulli}(\sigma(f_{\boldsymbol{\theta}}(\tilde{\mathbf{x}})))$, we define the classification analogue $PV_C(S)$ as the ratio of the epistemic and aleatoric components. Applying the same idealization as in the regression setting yields the measure
\[
IPV_{C}(S)
=
\frac{1}{N}
\mathbb{E}_{\tilde{\mathbf{x}} \sim F_{in}}
\left[
\bar{u}(\tilde{\mathbf{x}})
\,
\bar{g}_S(\tilde{\mathbf{x}})^{\top}
\left(
\Lambda^{C}_{S}
+
\frac{1}{N} V_S
\right)^{-1}
\bar{g}_S(\tilde{\mathbf{x}})
\right],
\]
where $\bar{u}(\tilde{\mathbf{x}})$ denotes the Bernoulli variance of $\tilde{y}$, and
\[
\Lambda^{C}
=
\mathbb{E}_{\tilde{\mathbf{x}} \sim F_{in}}
\left[
\bar{u}(\tilde{\mathbf{x}})
\,\bar{g}(\tilde{\mathbf{x}})
\bar{g}(\tilde{\mathbf{x}})^{\top}
\right].
\]
Using arguments very similar to those in the proof of Theorem~1 (with appropriate adjustments for $\bar{u}(\tilde{\mathbf{x}})$), we obtain the following result.

\paragraph{Theorem (Classification monotonicity).}
For subsets $S, S' \subseteq \{1,2,\cdots,p\}$, we have
\[
IPV_C(S) \le IPV_C(S')
\quad \text{whenever } S \subseteq S'.
\]

\subsection{Proof of Theorem 2} \label{proof:thm2}

\noindent
By (\ref{tridentity}), for any $S$ with $|S| = k$, we have 
 
\begin{eqnarray*}
\frac{N}{\sigma_0^2} IPV(S) 
&=& tr \left( \left( \Lambda_{S,S} + \frac{\sigma_0^2 V_{S,S}}{N} \right)^{-1} \Lambda_{S,S} \right)\\
&=& tr \left( \left( D_{S,S} + C_k + \tilde{c}I_k \right)^{-1} (D_{S,S} + C_k) \right)\\
&=& tr \left( \left( \tilde{c} (D_{S,S} + C_k)^{-1} + I_k \right)^{-1} \right)
\end{eqnarray*}

\noindent
where $\tilde{c} = \frac{c \sigma_0^2}{N}$ and $C_k$ is the common value taken by any $k$-dimensional principal sub-matrix of $C$ (due to permutation invariance). Let $m_j$ denote the index of the $j^{th}$ largest diagonal entry of $\Lambda$, for $1 \leq j \leq p$, and let 
 
\[
S^* \stackrel{\Delta}{=} \{m_1,m_2,\cdots, m_k\}. 
\]

\noindent
Note that $IPV(S)$, by construction, is invariant to any re-ordering of the elements in $S$. Let $\sigma:S^* \rightarrow S^*$ be an ordering of the elements of $S^*$ where $m_1$ appears at the first position, $m_2$ at the second position, and so on. By the permutation invariance of $C$, it follows that 
 
\[
\frac{N}{\sigma_0^2} IPV(S^*) = tr \left( \left( \tilde{c} (D_{\sigma(S^*),\sigma(S^*)} + C_k)^{-1} + I_k \right)^{-1} \right)
\]

\noindent
Now, note that by construction $D_{S,S} \preceq D_{\sigma(S^*),\sigma(S^*)}$ for any subset $S$ of size $k$ (here $\preceq$ denotes the Loewner ordering). Using the reversal of the Loewner ordering under inversion twice, we obtain

\[
\left( \tilde{c} (D_{S,S} + C_k)^{-1} + I_k \right)^{-1} \; \preceq \; \left( \tilde{c} (D_{\sigma(S^*),\sigma(S^*)} + C_k)^{-1} + I_k \right)^{-1}
\]

\noindent
It follows that

\begin{equation} \label{maxksubset}
IPV(S^*) = \max_{S: |S| = k} IPV(S). 
\end{equation}

Similarly, if $\tilde{m}_j$ denote the index of the $j^{th}$ smallest diagonal entry of $\Lambda$, for $1 \leq j \leq p$, and 
 
\[
\tilde{S}^* \stackrel{\Delta}{=} \{\tilde{m}_1,\tilde{m}_2,\cdots, \tilde{m}_k\}, 
\]

\noindent
then we get 
 
\begin{equation} \label{minksubset}
IPV(\tilde{S}^*) = \min_{S: |S| = k} IPV(S). 
\end{equation}

\noindent
We now show that, with high probability, the proposed \textit{Gradient--Laplace} algorithm selects $S^*$, whereas the \textit{Subnet Diagonal} algorithm of \cite{daxberger2021bayesian} selects $\tilde{S}^*$. This follows essentially from the fact that $\tilde{g}$ (see Algorithm \ref{alg:neuralnet2}) is close to $\operatorname{diag}(\Lambda)$. By permutation invariance of $C$, each diagonal entry of $\Lambda$ is obtained by adding a common non-negative constant (the shared diagonal entry of $C$) to the corresponding diagonal entry of $D$. Consequently, any algorithm that depends on the relative ordering of entries in $\tilde{g}$ should, in the limit, agree with one based on the relative ordering of entries in $D$.

To formalize the heuristic argument above, we require Assumptions (A1) and (A4) from Section~A.1. For simplicity, we will consider a version of Algorithm \ref{alg:neuralnet2} that uses the training data as the reference data. Using the bound in (\ref{HSbound}), it follows that 
 
\[
\sup_{1 \leq i \leq p} \left| \frac{1}{N} \sum_{n=1}^N {g}_i^2({\bf x}_n) - \Lambda_{ii} \right| = o_{P_{true}} \left( \sqrt{\frac{\log p}{N}} \right). 
\]

We need one additional mild assumption concerning the ``separation'' of the relevant diagonal entries of $D$ (or equivalently $\Lambda$). 

\begin{itemize}
    \item (A6) We assume either that every diagonal entry of $D$ that is strictly smaller than $D_{m_k m_k}$ differs from it by at least $\sqrt{\frac{\log p}{N}}$, or that $D_{m_k m_k}$ is the smallest diagonal entry of $D$ (i.e., the $u$-th largest diagonal entry of $D$ equals $D_{m_k m_k}$ for every $k < u \le p$). 
\end{itemize} 

\noindent
For example, this condition will be satisfied with any $k$ if $\min_{i,j: D_{ii} \neq D_{jj}} |D_{ii} - D_{jj}| > \sqrt{\frac{\log p}{N}}$. 

%
%

\noindent
Since the Gradient Laplace algorithm chooses the $k$ indices which maximize the value of $\sum_{n=1}^N {g}_i^2({\bf x}_n)$, on an event with $P_{true}$-probability tending to one, it will end up choosing the $k$ indices which maximize the value of $\Lambda_{ii}$ (or equivalently $D_{ii}$). It follows that 
 
\[
P_{true} \left( IPV(S_{GrL,k}) = IPV(\{m_1, m_2, \cdots, m_k\}) \right) \to 1. 
\]

\noindent
as $N \rightarrow \infty$. Combining this with (\ref{maxksubset}), we get 
 
\[
P_{true} \left( \text{Dis}(S_{GrL,k}) = \min_{S: |S| = k} \text{Dis}(S) \right) \rightarrow 1
\]

\noindent
as $N \rightarrow \infty$. Since the {\it Subnet diagonal} algorithm works by choosing the indices corresponding to the $k$ smallest entries of $\tilde{g}$, a repeat of the above argument focusing now on the $k$ smallest entries of $\tilde{g}$, and of the diagonals of $\Lambda$ and $D$ (along with a relevant separability condition) gives 
 
\[
P_{true} \left( \text{Dis}(S_{SubD,k}) = \max_{S: |S| = k} \text{Dis}(S) \right) \rightarrow 1, \quad \text{as}\quad  N \rightarrow \infty.
\]

\subsection{Proof of Theorem 3} \label{proof:thm3}

\noindent
We say that $\Lambda$ is $\epsilon$-diagonally dominant if 
 
    \[
    \epsilon \Lambda_{ii} > \sum_{j \neq i} |\Lambda_{ij}|
    \]

\noindent
for every $1 \leq i \leq p$. Recall from the proof of Theorem \ref{thm2} that $m_j$ denotes the index of the $j^{th}$ largest diagonal entry of $\Lambda$, for $1 \leq j \leq p$, and 
 
\[
S^* \stackrel{\Delta}{=} \{m_1,m_2,\cdots, m_k\}. 
\]

\noindent
Similarly, $\tilde{m}_j$ denotes the index of the $j^{th}$ smallest diagonal entry of $\Lambda$, for $1 \leq j \leq p$, and 
 
\[
\tilde{S}^* \stackrel{\Delta}{=} \{\tilde{m}_1,\tilde{m}_2,\cdots, \tilde{m}_k\}. 
\]

\noindent
For simplicity, we will denote $\Lambda_{S^*, S^*}$ by $\Lambda_1$ and $\Lambda_{\tilde{S}^*, \tilde{S}^*}$ by $\Lambda_2$. Let $D_1$ denote the diagonal matrix consisting of the diagonal entries of $\Lambda_1$, and $E_1 = \Lambda_1 - D_1$. Similarly, define $D_2$ and $E_2$ based on $\Lambda_2$. Note by the $\epsilon$-diagonal dominance of $\Lambda$ that $\epsilon D_i + E_i$ and $\epsilon D_i - E_i$ are diagonally dominant matrices for $i = 1,2$. Hence

\begin{eqnarray*}
\Lambda_1 - \Lambda_2 
&=& D_1 - D_2 + E_1 - E_2\\
&=& D_1 - D_2 - \epsilon D_1 - \epsilon D_2 + (\epsilon D_1 + E_1) + (\epsilon D_2 - E_2)\\
&\succ& D_1 - D_2 -\epsilon D_1 - \epsilon D_2\\
&=& (1-\epsilon) D_1 - (1+\epsilon) D_2\\
&\succ& (1 - \epsilon) \Lambda_{m_k m_k} I_k - (1+\epsilon) \Lambda_{m_{p-k+1} m_{p-k+1}} I_k\\
&\succ& 0. 
\end{eqnarray*}

\noindent
The last inequality follows by the ratio condition for $\Lambda_{m_k m_k}$ and $\Lambda_{m_{p-k+1} m_{p-k+1}}$. Hence, for any constant $c > 0$, it follows that 
 
\[
tr \left( (\Lambda_1 + cI_k)^{-1} \right) \leq tr \left( (\Lambda_2 + cI_k)^{-1} \right). 
\]

\noindent
Since $V$ is a multiple of identity, and 
 
    \[
    (\Lambda_i + cI_k)^{-1} \Lambda_i = I_k - c (\Lambda_i + cI_k)^{-1} 
    \]

\noindent
for $i = 1,2$, it follows that 
 
    \[
    IPV(S^*) \geq IPV(\tilde{S}^*). 
    \]

\noindent
Instead, if the ratio condition holds for $\Lambda_{m_k m_k}$ and $\Lambda_{m_{k+1} m_{k+1}}$, then by similar arguments as above, it follows that 
 
    \[
    IPV(S^*) \geq IPV(S) 
    \]

\noindent
for any $S$ with $S \cap S^* = \emptyset$. The result follows by noting, exactly as in the proof Theorem \ref{thm2}, that 
 
\[
P_{true} \left( \text{Dis}(S_{GrL,k}) = \min_{S: |S| = k} \text{Dis}(S) \right) \rightarrow 1
\]

\noindent
and 
 
\[
P_{true} \left( \text{Dis}(S_{SubD,k}) = \max_{S: |S| = k} \text{Dis}(S) \right) \rightarrow 1 
\]

\noindent
as $N \rightarrow \infty$.

\section{Benchmark Algorithms}\label{app:benchmark_algo} We compare the proposed methods with the following benchmark algorithms.
\begin{itemize}
    \item \textbf{NeuralLinear}~\citep{snoek2015scalable,riquelme2018deep} is a sub-network LA based approach where the subset of indices for the surrogate $S$ is chosen to be the indices associated with the last layer of the parameters of the network.
    
       \item \textbf{Last $k$} is the natural $k$-dependent extension of NeuralLinear. We order parameters from the first layer to the last layer and choose the final $k$ parameters as the subset $S$. This baseline tests whether simply concentrating the Laplace approximation near the output layer is sufficient.
       
    \item \textbf{Subnet diagonal}~\citep{daxberger2021bayesian} is another sub-network LA approach, which for a given subset size $k$, selects the indices associated to the $k$ smallest diagonal entries of $\Omega$ as the elements of $S$.

    \item \textbf{MAP} is a deterministic baseline used in the Wheel Bandit experiment. It selects actions using the point estimate $f_{\hat{\bm{\theta}}}$ without posterior sampling, and therefore performs no uncertainty-driven exploration.

    \item \textbf{Deep Ensemble} method~\citep{lakshminarayanan2017simple,NIPS2016_8d8818c8} generates $B$ bootstrap samples using the training data, uses them to train the network separately, and then proposes the $2.5\%$ and $97.5\%$ quantiles to be the lower and upper limit respectively of a $95\%$ confidence interval for $f_{{\tht}}(\mathbf{x})$.  

    \item \textbf{KFAC-Laplace}~\citep{ritter2018scalable,martens2015optimizing} uses a Kronecker-factored block approximation to the Laplace precision matrix. It is a structured full-network baseline: unlike sub-network methods, it does not select a subset $S$, but approximates curvature layer-wise using Kronecker factors. In our plots it is therefore shown as a $k$-independent reference.

\end{itemize}

\section{Additional Experimental Details for Section~\ref{subsec:predictive_approx_accuracy}}
\label{app:wass_real}

This appendix provides the full experimental details for the two real-world setups summarized in Section~\ref{subsec:predictive_approx_accuracy}: \textbf{Setup A}, the YEAR Prediction MSD regression task, and \textbf{Setup B}, the Binary CIFAR-10 classification task. Together they exercise the regression and binary-classification specializations of our predictive-variance metric, on backbones with $p = 98{,}801$ parameters for YEAR and $p = 1{,}730{,}129$ parameters for Binary CIFAR-10.

\subsection{Setup A: YEAR Prediction MSD (Regression)}
\label{app:wass_real:year}

\paragraph{Data.} We use the \emph{YearPredictionMSD} dataset~\citep{bertinmahieux2011yearpredictionmsd} from the UCI repository, with the canonical $463{,}715/51{,}630$ train/test split shipped with the dataset. The release-year target $y$ is standardized to zero-mean unit-variance using the training partition only; the same scaler is applied to the test set. Each input $\mathbf{x} \in \mathbb{R}^{90}$ is the $90$-dimensional timbre summary statistic ($12$ timbre averages and $78$ timbre covariances). No further feature engineering is performed.

\paragraph{Working model.} The working model is a fully connected feed-forward neural network with three hidden layers of $200$ ReLU units and a scalar linear output head:
\[
    f_{\bm{\theta}}(\mathbf{x}) = W_{4}\,\mathrm{ReLU}\!\left(W_{3}\,\mathrm{ReLU}\!\left(W_{2}\,\mathrm{ReLU}(W_{1}\mathbf{x} + b_{1}) + b_{2}\right) + b_{3}\right) + b_{4},
\]
giving $p=98{,}801$ trainable parameters.

\paragraph{Training.} We train with the mean squared error loss using Adam~\citep{kingma2014adam} (initial learning rate $10^{-3}$, default $(\beta_{1}, \beta_{2}) = (0.9, 0.999)$, batch size $256$) for $100$ epochs, under a cosine annealing learning-rate schedule with $T_{\max} = 100$ and $\eta_{\min} = 0$. We impose IID $N(0, 1)$ priors on the weights and biases, fix the prior precision at $\alpha = 1$, and use the residual mean squared error of the MAP estimate as our plug-in estimator for $\sigma_{0}^{2}$ in~\eqref{Gauss:Newton:precision}, floored at $10^{-3}$ for numerical stability. 

\paragraph{Evaluation.} For each method and subset size $k$ in  $\{50, 100, 200, 500, 1000, 2000, 5000, 10{,}000\}$, we compute the per-test-point Wasserstein distance $|\sigma_{\mathrm{Full}}(\mathbf{x}^{*}) - \sigma_{\mathrm{method}}(\mathbf{x}^{*})|$ and average it over the first $10{,}000$ rows of the canonical YEAR test split. We then average across ten independent random seeds and report the mean and 95\% band in Figure~\ref{fig:wass_year}. 

\paragraph{Oracle for the coverage panel.} The right panel of Figure~\ref{fig:wass_year} reports the empirical $95\%$-credible-interval coverage of the regression target. For the secondary coverage diagnostic in Figure~\ref{fig:wass_year}, the oracle target $f_{\bm{\theta}_{0}}(\mathbf{x}^{*})$ is approximated using the predictions of a high-capacity reference model trained on the full $463{,}715$-row training partition with the same architecture but a longer schedule; this oracle achieves a held-out test mean squared error consistent with the published baselines for this dataset and is used only on the test set, never in the working model's training loop. The coverage indicator at the $95\%$ level is the empirical mean of $\mathbf{1}\{|f_{\mathrm{oracle}}(\mathbf{x}^{*}) - f_{\mathrm{MAP}}(\mathbf{x}^{*})| \leq 1.96 \cdot \sigma_{\mathrm{method}}(\mathbf{x}^{*})\}$ over the test subsample; this is the (epistemic-only) credible-interval form of the coverage metric.

\paragraph{Deep-Ensemble overlay.} The horizontal dash-dotted line in the right panel of Figure~\ref{fig:wass_year} reports the empirical coverage of the same $95\%$ credible interval for an $M$-member \emph{Deep Ensemble}~\citep{lakshminarayanan2017simple}, with the ensemble predictive mean $f_{\mathrm{ens}}(\mathbf{x}^{*}) = \frac{1}{M} \sum_{m} f_{\bm{\theta}_{m}}(\mathbf{x}^{*})$ and the (unbiased) cross-member variance as the predictive variance $\sigma_{\mathrm{ens}}^{2}(\mathbf{x}^{*})$. The ensemble members reuse the architecture, recipe, $n_{\mathrm{train}}$, and seed-set of the LA suite; see the Working model section.

\subsection{Setup B: Binary CIFAR-10 (Classification)}
\label{app:wass_real:cifar_binary}

\paragraph{Data and binary reduction.} We use the standard CIFAR-10 dataset~\citep{krizhevsky2009learning} with the canonical class indices $0, 1, \dots, 9$. We construct a binary task by collapsing the ten classes into two balanced groups of five via the the simple rule $y_{\mathrm{binary}} = \mathbf{1}\{y \geq 5\}$. Under this, classes $\{$\texttt{airplane}, \texttt{automobile}, \texttt{bird}, \texttt{cat}, \texttt{deer}$\}$ map to group $0$ and $\{$\texttt{dog}, \texttt{frog}, \texttt{horse}, \texttt{ship}, \texttt{truck}$\}$ map to group $1$. 
The split is class-balanced, with $25{,}000$ training images per group, but otherwise arbitrary. We also tested a semantic animal-versus-vehicle split and obtained qualitatively similar results, indicating that our conclusions are largely agnostic to the particular binary reduction used, rather than being specific to the threshold split $y_{\mathrm{binary}}=\mathbf{1}\{y\geq 5\}$.

\paragraph{Working model.} The working model is a CIFAR-style ResNet~\citep{he2016deep}, configured as three stages of $\{16, 32, 64\}$ filters with $n$ residual basic-blocks per stage ($n \in \{3, 5, 9, 18\}$ for ResNet-20 / 32 / 56 / 110 respectively) and global average pooling before the classification head. The standard $\mathrm{Linear}(64, 10)$ classification head is replaced by a single-output $\mathrm{Linear}(64, 1)$ logit. The single output $f_{\bm{\theta}}(\mathbf{x})$ is interpreted as the binary log-odds, with $P(y_{\mathrm{binary}} = 1 \mid \mathbf{x}, \bm{\theta}) = \sigma(f_{\bm{\theta}}(\mathbf{x}))$. Head-swapped trainable parameter counts are $p = 271{,}889$ (ResNet-20), $466{,}906$ (ResNet-32), $855{,}930$ (ResNet-56), and $1{,}730{,}129$ (ResNet-110); the headline figure in Section~\ref{subsec:predictive_approx_accuracy} (Figure~\ref{fig:wass_cifar_binary}) uses ResNet-110, and the smaller backbones are reported in Section~\ref{app:wass_real:scaling} below.

\paragraph{Training.} We train with the binary cross-entropy (BCE) loss on $f_{\bm{\theta}}(\mathbf{x})$ using SGD with Nesterov momentum $0.9$, weight decay $10^{-4}$, initial learning rate $0.1$, and batch size $128$, under a cosine annealing learning-rate schedule with $T_{\max} = 100$ epochs and $\eta_{\min} = 0$. Standard CIFAR-10 augmentation is applied during training (random crops with $4$-pixel reflective padding and random horizontal flips), with channel-wise standardization to the canonical CIFAR-10 mean and standard deviation. The augmentation is disabled for the Wasserstein subsample, the test set, and all Jacobian computations to keep these quantities deterministic across replications. Just as in Appendix~\ref{app:wass_real:year}, we impose IID $N(0, 1)$ priors on the weights and biases and fix the prior precision at $\alpha = 1$.

%
%

\paragraph{Evaluation.} For each method and subset size $k \in \{50, 100, 200, 500, 1000, 2000, 5000, 10{,}000\}$, we compute the per-test-point Wasserstein distance $|\sigma_{\mathrm{Full}}(\mathbf{x}^{*}) - \sigma_{\mathrm{method}}(\mathbf{x}^{*})|$ and average it over a fixed $1{,}000$-image test subsample drawn deterministically per replication. As in Setup A, the Gauss--Newton precision matrix is built on a uniformly subsampled training subset of $N = 2{,}000$ images per replication; this subsample is drawn afresh from the full $50{,}000$-row training set per seed. We average over ten independent seeds and report the mean and $1.96 \times $SD confidence bars in Figure~\ref{fig:wass_cifar_binary}.

\paragraph{Oracle for the coverage panel.} The right panel of Figure~\ref{fig:wass_cifar_binary} reports the empirical $95\%$-credible-interval coverage of the binary log-odds. Since training a separate full-data oracle for every binary partition of CIFAR-10 is wasteful, we instead reuse a single high-capacity multiclass oracle and collapse it to binary log-odds at evaluation time. Specifically, we take a wide ResNet trained on the full 10-class CIFAR-10 task to high accuracy (the WRN-28-10 RobustBench standard model~\citep{croce2020robustbench} is used in our runs, with $94.77\%$ multi-class test accuracy on the $10{,}000$-image test split), apply the soft-max to its logits, and compute the binary log-odds
 
\begin{align}\label{eq:oracle_logodds}
    f_{\mathrm{oracle}}(\mathbf{x}^{*}) \;=\; \log \sum_{c \geq 5} p_{\mathrm{oracle}}(c \mid \mathbf{x}^{*}) - \log \sum_{c < 5} p_{\mathrm{oracle}}(c \mid \mathbf{x}^{*}).
\end{align}
 
This is the natural binary analogue of ``the oracle's preferred-class logit'' --- it matches the single-logit output of our working model in scale and sign, and is computed entirely on the test partition. The coverage indicator at the $95\%$ level is then the empirical mean of $\mathbf{1}\{|f_{\mathrm{oracle}}(\mathbf{x}^{*}) - f_{\mathrm{MAP}}(\mathbf{x}^{*})| \leq 1.96 \cdot \sigma_{\mathrm{method}}(\mathbf{x}^{*})\}$ over the test subsample, where $\sigma_{\mathrm{method}}(\mathbf{x}^{*})$ is the predictive standard deviation under~\eqref{posterior:predictive:variance} with $\Omega$ replaced by the relevant subnetwork surrogate of $\Omega$ in~\eqref{Gen_Gauss:Newton:precision}.

\subsection{Implementation and Reproducibility}
\label{app:wass_real:repro}

\paragraph{Computational details.} All experiments are run on a single NVIDIA H100 ($80$\,GB) per replication. The per-sample Jacobian $J(\mathbf{x}_{n})$ is computed in chunks via the \texttt{torch.func.vmap} composition $\texttt{vmap}(\texttt{grad}(\cdot))$ applied to a single-input forward pass, with chunk size $128$ for ResNet-20 / 32, $64$ for ResNet-56, and $32$ for the headline ResNet-110, chosen so that the per-chunk activation peak fits comfortably within H100 memory at every backbone scale. The $\Omega^{-1} J(\mathbf{x}^{*})$ inner product is computed using the Woodbury identity on the $N \times N$ Gram kernel of the training Jacobian, which keeps the maximal allocation at $4 N^{2}$ bytes ($\approx 16$\,MB at $N = 2{,}000$) regardless of $k$. The greedy hybrid step of \emph{Greedy-Laplace} (Algorithm~\ref{alg:neuralnet1}) operates on the $\bar{k} \times \bar{k}$ candidate sub-precision with $\bar{k} = \min(2k + 1{,}000, p - 1)$, capped at $\bar{k} = 30{,}000$ to bound the sub-precision allocation at $\approx 3.6$\,GB.

\paragraph{Method set and selection rules.} The methods evaluated in both setups are \emph{Gradient-Laplace}, \emph{Greedy-Laplace}, \emph{Subnet Diagonal}, \emph{Last $k$}, and \emph{NeuralLinear}; full definitions are in Appendix~\ref{app:benchmark_algo}. For the binary CIFAR-10 setup, the single-logit Bernoulli likelihood gives the Gauss--Newton precision
\[
\Omega
=
\sum_{n=1}^{N}
p_n(1-p_n)\,
g(\mathbf{x}_n)g(\mathbf{x}_n)^\top
+
\alpha I,
\qquad
p_n=\sigma(f_{\bm{\theta}}(\mathbf{x}_n)).
\]
Therefore the diagonal score used by \emph{Gradient-Laplace} and \emph{Subnet Diagonal} is
\[
\mathrm{diag}(\Omega)_i
=
\sum_{n=1}^{N}
p_n(1-p_n)\,
g_i(\mathbf{x}_n)^2
+
\alpha .
\]
For the regression setup, the corresponding score is
\[
\mathrm{diag}(\Omega)_i
=
\sigma_0^{-2}
\sum_{n=1}^{N}
g_i(\mathbf{x}_n)^2
+
\alpha .
\]


\emph{NeuralLinear} restricts $S$ to the parameters of the final linear layer ($k = 201$ for the YEAR MLP head and $k = 65$ for the binary CIFAR-10 head); it has no $k$-dependence and appears as a single hollow-diamond marker on each panel of Figures~\ref{fig:wass_year} and~\ref{fig:wass_cifar_binary}. \emph{Last $k$} takes the trailing $k$ indices of the flat parameter vector. 

\paragraph{Replications and reporting.} Each setup uses ten independent random seeds; so the seed-to-seed variability captured by the error bars in Figures~\ref{fig:wass_year} and~\ref{fig:wass_cifar_binary} reflects independent training runs. All seeds and the figure-rendering scripts are released with the public code accompanying this paper.

\subsection{Scaling: Smaller-Backbone Companions to ResNet-110}
\label{app:wass_real:scaling}
To check that the Setup B conclusions are not specific to the ResNet-110 backbone reported in the main text, we ran the same Binary CIFAR-10 experiment across the CIFAR-style ResNet family: ResNet-20, ResNet-32, and ResNet-56. We used the same training recipe, $k$-grid, and method set across all backbones. The qualitative ordering of the methods is stable across this sweep: \emph{Gradient-Laplace} and \emph{Greedy-Laplace} consistently provide substantially closer approximations to the full Laplace predictive distribution than \emph{Subnet Diagonal} and \emph{Last $k$}. 


\begin{figure}[h]
    \centering
    \includegraphics[width=\linewidth]{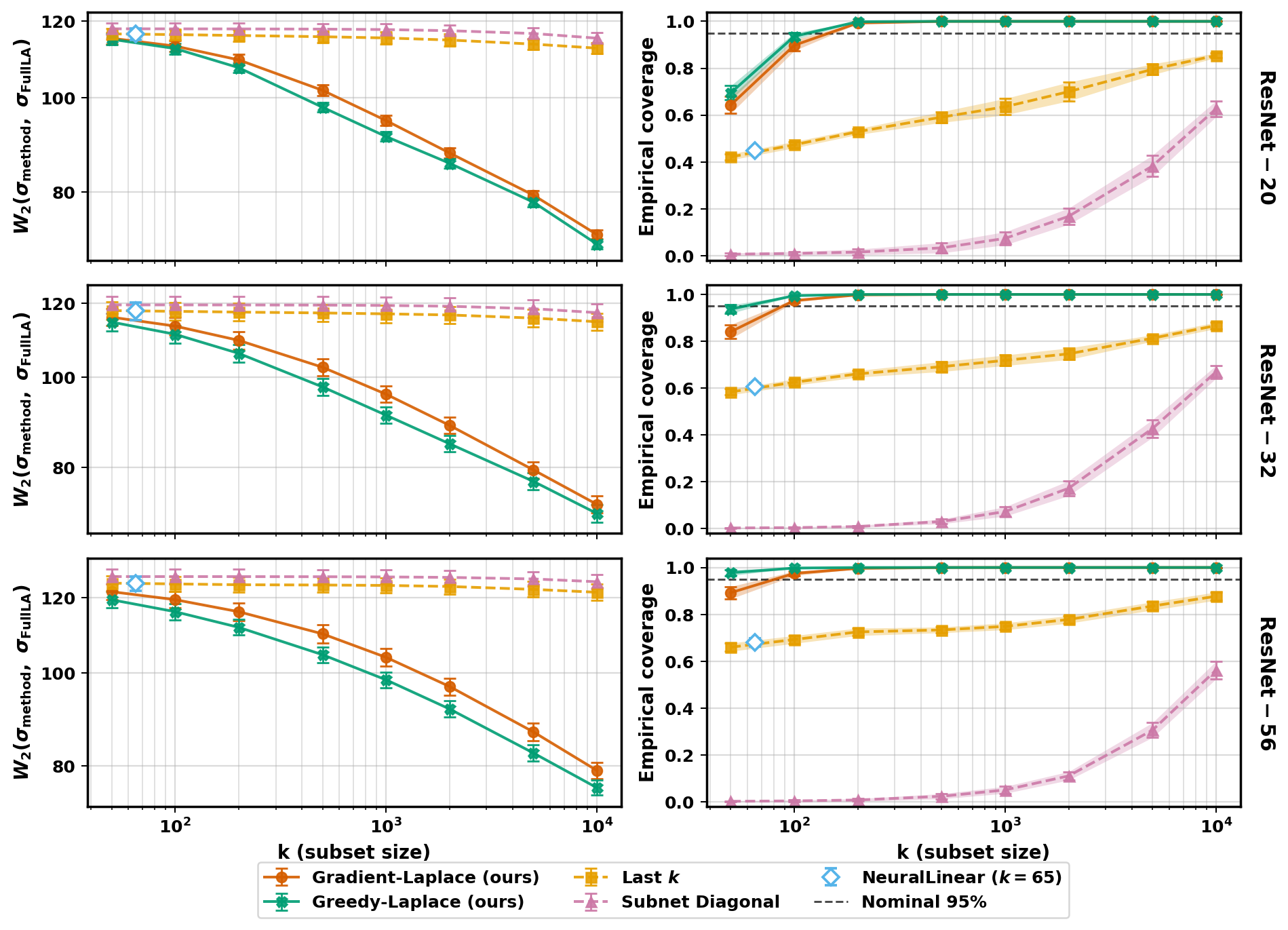}
    \caption{Setup B at the three smaller CIFAR-style ResNet backbones. \textbf{Top:} ResNet-20 ($p = 271{,}889$). \textbf{Middle:} ResNet-32 ($p = 466{,}906$). \textbf{Bottom:} ResNet-56 ($p = 855{,}930$). All three sweeps use the same training recipe, $k$-grid, method set, and ten random seeds as the ResNet-110 headline in Figure~\ref{fig:wass_cifar_binary}; layout, axes, and legend are identical. The gap between the proposed methods and the diagonal baselines on the $W_{2}$ panel is roughly an order of magnitude across the full $k$-grid at every backbone. The advantage of \emph{Greedy-Laplace} over \emph{Gradient-Laplace} is essentially zero at ResNet-20 (curves overlap within seed noise) and opens up gradually as $p$ grows toward the headline ResNet-110.}
    \label{fig:wass_cifar_binary_scaling}
\end{figure}

\subsection{Setup C: Multi-class CIFAR-10 with the Proper Softmax Hessian}
\label{app:wass_real:cifar_classification}

The binary reduction in Setup B is a clean instance of the scalar Hessian factor $p(1 - p)$ that appears in the binary-classification specialization of~\eqref{Gen_Gauss:Newton:precision}. To verify that the qualitative ordering of the proposed methods is not an artifact of this scalar reduction, we additionally report results on the full ten-class CIFAR-10 task.

\paragraph{Precision matrix and predictive distribution.} Let $\mathbf{p}_{n} = \mathrm{softmax}(g_{\hat{\bm{\theta}}}(\mathbf{x}_{n})) \in \mathbb{R}^{C}$ denote the MAP soft-max probability vector at training input $\mathbf{x}_{n}$. The softmax Hessian is
\[
    H_{\mathrm{softmax}}(\mathbf{x}_{n}) = \mathrm{diag}(\mathbf{p}_{n}) - \mathbf{p}_{n}\mathbf{p}_{n}^{\top} \;+\; \varepsilon I_{C},
\]
where $\varepsilon = 10^{-6}$ is a small numerical ridge that lifts the rank-deficient $\mathrm{diag}(\mathbf{p}_{n}) - \mathbf{p}_{n}\mathbf{p}_{n}^{\top}$ off the all-ones null direction. Eigendecomposing $H_{\mathrm{softmax}}(\mathbf{x}_{n}) = U_{n} \mathrm{diag}(\bm{\lambda}_{n}) U_{n}^{\top}$ and defining $W_{n} = U_{n}\mathrm{diag}(\sqrt{\bm{\lambda}_{n}})$ gives $H_{\mathrm{softmax}} = W_{n} W_{n}^{\top}$, and stacking the row-rescaled Jacobians $W_{n}^{\top} J(\mathbf{x}_{n}) \in \mathbb{R}^{C \times p}$ across the $N$ training rows produces $J_{w}^{\mathrm{full}} \in \mathbb{R}^{NC \times p}$ such that the precision in~\eqref{Gen_Gauss:Newton:precision} reduces to $\Omega = (J_{w}^{\mathrm{full}})^{\top} J_{w}^{\mathrm{full}} + \alpha I_{p}$. 

The predictive distribution at a test input $\mathbf{x}^{*}$ is now $C$-dimensional: $f(\mathbf{x}^{*}) \sim N(g_{\mathrm{MAP}}(\mathbf{x}^{*}), \Sigma_{\mathrm{pred}}(\mathbf{x}^{*}))$ with $\Sigma_{\mathrm{pred}}(\mathbf{x}^{*}) = J(\mathbf{x}^{*}) \Omega^{-1} J(\mathbf{x}^{*})^{\top} \in \mathbb{R}^{C \times C}$. Since the $C \times C$ posterior covariance is too rich to compare full-bandwidth against subnetwork surrogates pointwise, we extract per-class diagonals $\sigma_{c}(\mathbf{x}^{*}) = \sqrt{\Sigma_{\mathrm{pred}}(\mathbf{x}^{*})[c, c]}$ and define the multi-class Wasserstein metric as
\[
    W_{2}(\mathbf{x}^{*}) \;=\; \frac{1}{C} \sum_{c = 1}^{C} \left| \sigma_{\mathrm{Full}}(\mathbf{x}^{*}, c) - \sigma_{\mathrm{method}}(\mathbf{x}^{*}, c) \right|.
\]

\paragraph{Working models.} We use the same CIFAR-style ResNet family as in Setup B, retaining the standard $\mathrm{Linear}(64, 10)$ classification head, and we train with the cross-entropy loss for $100$ epochs under the same SGD recipe (lr $= 0.1$, Nesterov momentum $0.9$, weight decay $10^{-4}$, cosine annealing). We report results for ResNet-20 ($p = 272{,}474$, $N = 2{,}000$ training rows) and ResNet-32 ($p = 466{,}906$, $N = 500$ training rows --- the multi-output Jacobian at $N = 2{,}000$, $C = 10$, $p = 466{,}906$ exceeds the $80$\,GB H100 budget once the Cholesky and predictive-variance scratch are also resident). ResNet-56 and ResNet-110 exceed the practical single-H100 budget used in our implementation under the full multi-class Gauss--Newton formulation. One scalable and often used alternative is to approximate the multi-class posterior by treating the ten class-logit heads separately, thereby ignoring cross-class Hessian terms; this reduces memory and computation substantially and would extend the comparison to larger backbones, but we leave this approximation outside the present evaluation.


\paragraph{Evaluation.} For each backbone and subset size $k$ in $\{50, 100, 200, 500, 1000, 2000, 5000, 10{,}000\}$, we compute the per-test-point class-averaged Wasserstein metric on a fixed $1{,}000$-image test subsample, and average over ten independent random seeds. Coverage on the right panel uses the same WRN-28-10 oracle as in Setup B but applied directly at the multi-class level: the indicator at the $95\%$ level is $\mathbf{1}\{|f_{\mathrm{oracle}, c}(\mathbf{x}^{*}) - f_{\mathrm{MAP}, c}(\mathbf{x}^{*})| \leq 1.96 \cdot \sigma_{\mathrm{method}}(\mathbf{x}^{*}, c)\}$, averaged over both the test points and the ten classes (i.e.\ the per-class-marginal coverage convention used for this secondary diagnostic).

\begin{figure}[h]
    \centering
    \includegraphics[width=\linewidth]{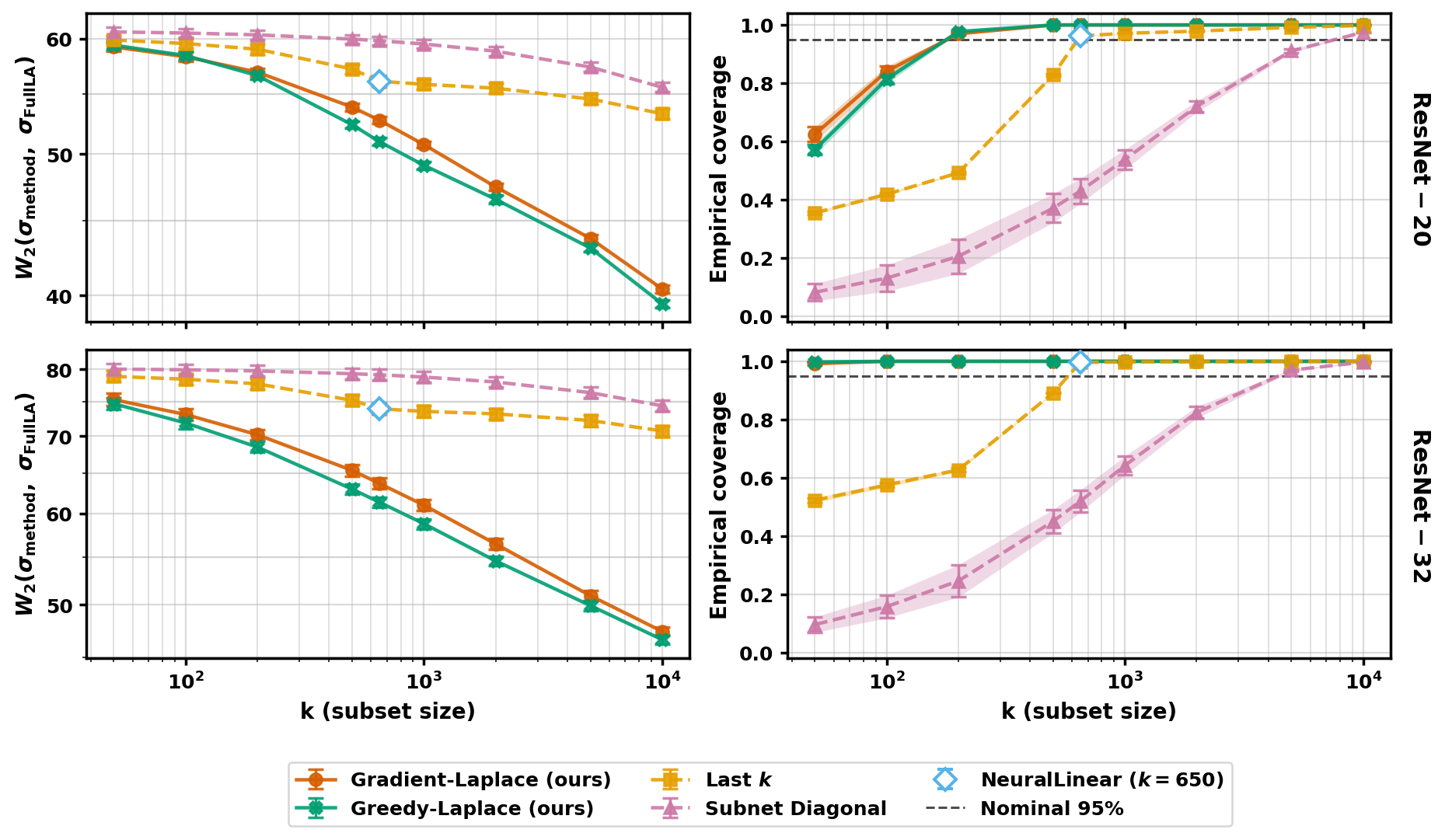}
    \caption{Setup C (multi-class CIFAR-10, proper softmax-Hessian formulation). \textbf{Top:} ResNet-20, $p = 272{,}474$, \textbf{Bottom:} ResNet-32, $p = 467{,}556$. The qualitative ordering matches Setups A and B; \emph{NeuralLinear} sits at $k = 650$ (the ten-output head). $W_{2}$ in this multi-class form is the per-class-averaged predictive standard-deviation gap defined above.}
    \label{fig:wass_cifar_classification}
\end{figure}

\paragraph{Summary of findings.} As shown in Figure~\ref{fig:wass_cifar_classification}, the qualitative ordering of the methods on Setup C reproduces that of Setups A and B: the proposed \emph{Gradient-Laplace} and \emph{Greedy-Laplace} methods remain co-best across the full $k$ grid, with $W_{2}$ decreasing approximately monotonically as $k$ grows from $50$ to $10{,}000$. \emph{Subnet Diagonal} and \emph{Last $k$} remain near their large-$k$ asymptotes throughout, and \emph{NeuralLinear} at $k = 650$ is matched by the proposed methods at substantially smaller $k$. 

\subsection{Setup D: UCI Tabular Regression}
\label{app:wass_real:uci}

To complement the large-scale YEAR regression benchmark and the CIFAR-10 image-classification setups, we also evaluate on four standard UCI regression datasets commonly used in Bayesian neural-network benchmarks~\citep{hernandez2015probabilistic}: \texttt{Boston} ($N=506$, $d=13$), \texttt{concrete} ($N=1030$, $d=8$), \texttt{energy} ($N=768$, $d=8$), and \texttt{wine} ($N=1599$, $d=11$). We use the standard train/test split convention from this benchmark suite. Features and targets are standardized to zero mean and unit variance using the training partition only.

\paragraph{Working models.}
We evaluate two fully connected ReLU MLP backbones. The \textbf{small MLP} has two hidden layers of width $50$ and a scalar linear output head, giving $p=50d+2{,}651$ trainable parameters. The \textbf{large MLP} has three hidden layers of width $200$ and a scalar linear output head, giving $p=200d+80{,}801$ trainable parameters. Both models are trained with mean squared error loss using Adam with initial learning rate $10^{-2}$ and cosine annealing over $1{,}500$ epochs. We use the test-set residual mean squared error of the MAP estimate as the plug-in estimator for $\sigma_0^2$, and just as in other setups impose independent $N(0,1)$ priors on all weights and biases, corresponding to prior precision $\alpha=1$.

\begin{figure}[h]
    \centering
    \includegraphics[width=\linewidth]{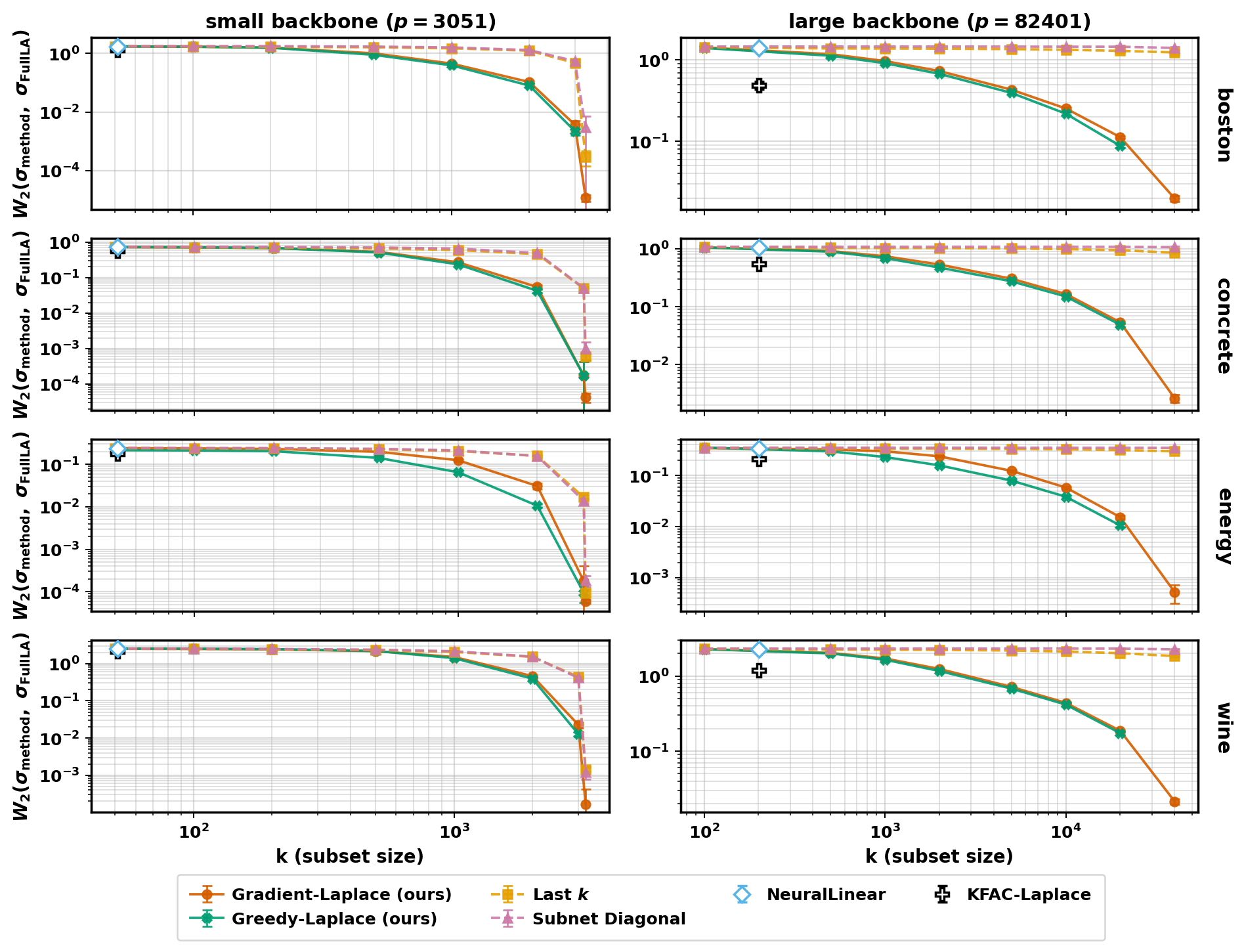}
    \caption{Setup D: UCI tabular regression. Average per-test-point Wasserstein distance between the full Laplace predictive distribution and each surrogate, across four UCI regression datasets. Rows correspond to datasets; columns correspond to the small and large MLP backbones. Lower is better.}
    \label{fig:wass_uci}
\end{figure}

\paragraph{Replications and reporting.}
Each dataset--backbone pair is evaluated over multiple random train/test splits and model initializations, with $50$ replications. We report the resulting Wasserstein metric in Figure~\ref{fig:wass_uci}.

\paragraph{Summary of findings.}
Across the four datasets and both backbone sizes, the qualitative ordering matches Setups A--C. \emph{Gradient-Laplace} and \emph{Greedy-Laplace} consistently produce closer approximations to the full Laplace predictive distribution than \emph{Subnet Diagonal}, \emph{Last $k$}, \emph{NeuralLinear}. \emph{KFAC-Laplace} provides a stronger structured baseline than diagonal-only approximations, but  is matched or surpassed by the proposed methods at moderate subset sizes. 

\section{Additional Experimental Details for Section~\ref{subsec:wheel_bandit}}
\label{app:wheel_bandit}

\subsection{Wheel Bandit Environment}

We use the Wheel Bandit environment of~\citet{riquelme2018deep}. The context $\mathbf{x}\in\mathbb{R}^{2}$ is sampled uniformly from the unit disk. There are five arms: one central arm and four outer arms (one for each quadrant). The central arm always has reward distribution $\mathcal{N}(\mu_c,\sigma_0^2)$. The outer arm corresponding to the quadrant containing $\mathbf{x}$ has reward distribution $\mathcal{N}(\mu_h,\sigma_0^2)$ when $\|\mathbf{x}\|>\delta$, and $\mathcal{N}(\mu_c,\sigma_0^2)$ otherwise. The remaining outer arms always have reward distribution $\mathcal{N}(\mu_c,\sigma_0^2)$.

We use $\mu_c=1.0$, $\mu_h=50$, and $\sigma_0=0.01$. The sparsity parameter is set to $\delta=0.95$, so the high-reward region occupies only $1-\delta^2=9.75\%$ of the disk.

\subsection{Training and Hyperparameter Details}

Each run begins with three random pulls per arm, for a total of $15$ warm-start interactions. After this warm start, the agent alternates between $20$ environment-interaction steps and $100$ stochastic-gradient updates on the replay buffer. The horizon is $T=16{,}000$, giving $800$ training phases.

The reward model is a fully connected neural network with two hidden layers of $100$ ReLU units and a scalar output. The input is the context concatenated with a one-hot arm encoding, giving input dimension $7$ and total parameter count $p=11{,}001$. We train with mean squared error loss using Adam with standard hyperparameters, learning rate $3\times 10^{-3}$, gradient clipping at $1.0$, and replay batch size $512$.

For Laplace-based methods, the prior precision is set to $\alpha=1$. The observation-noise variance used in the posterior is estimated online from the most recent $200$ residuals.


For a test context $\mathbf{x}^{*}$ and arm $a$, let
\[
g_S(\mathbf{x}^{*},a)
=
\nabla_{\bm{\theta}} f_{\bm{\theta}}(\mathbf{x}^{*},a)\big|_S
\]
denote the gradient restricted to the selected sub-network. The predictive variance used for Thompson sampling is
\[
\mathrm{Var}[\hat r(\mathbf{x}^{*},a)]
=
\sigma_0^2
+
g_S(\mathbf{x}^{*},a)^\top
\Omega_{S,S}^{-1}
g_S(\mathbf{x}^{*},a),
\]
where
\[
\Omega_{S,S}
=
\frac{1}{\sigma_0^2}J_S^\top J_S+\alpha I_k .
\]

\subsection{Cumulative Regret Traces and Detailed Results}

\begin{figure}[h]
    \centering
    \includegraphics[width=0.95\linewidth]{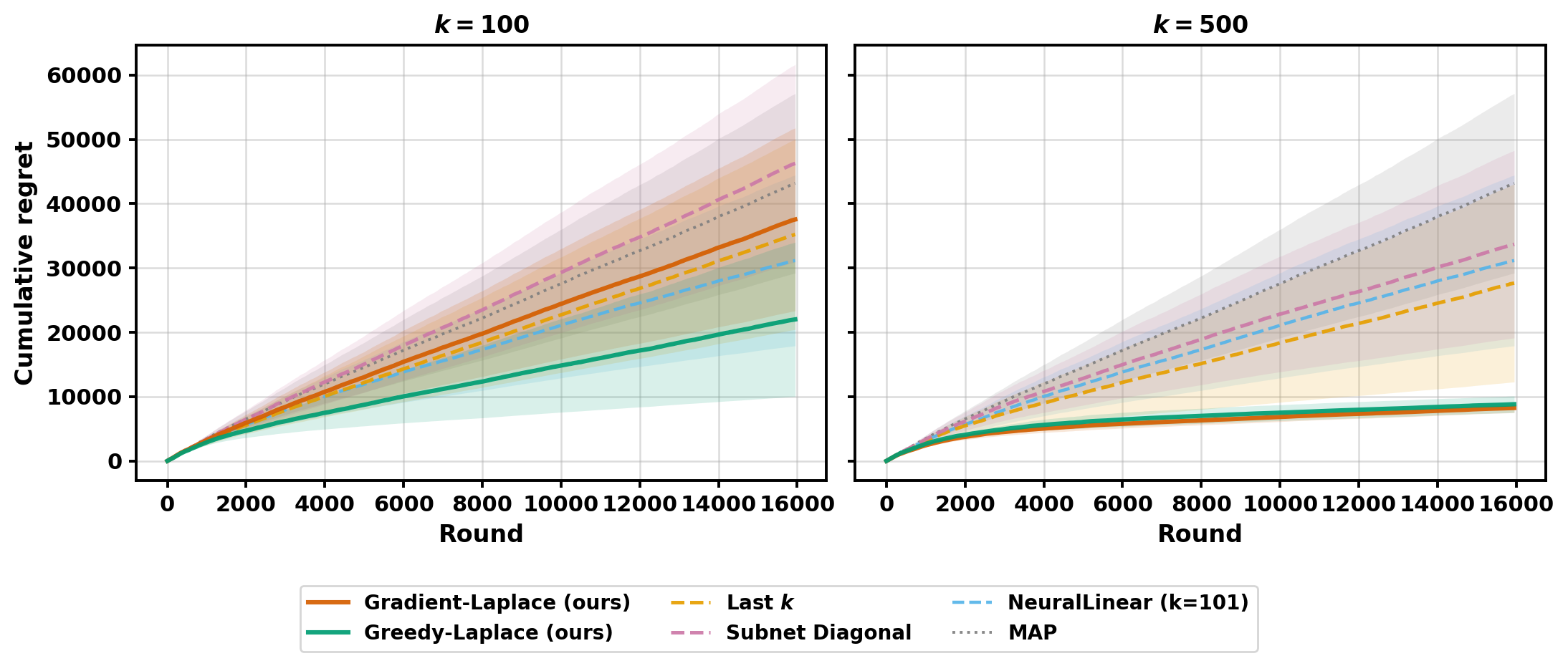}
    \caption{Cumulative regret on the Wheel Bandit at $\delta=0.95$. Lines show mean cumulative regret over twenty random seeds, with bands equal to $\pm 1.96\times \mathrm{SE}$. 
    }
    \label{fig:wheel_trace}
\end{figure}

Figure~\ref{fig:wheel_trace} reports cumulative regret over time, averaged over twenty random seeds.

Table~\ref{tab:creg_wheel} reports final cumulative regret at $T=16{,}000$.

\begin{table}[h]
\caption{Final cumulative regret on the Wheel Bandit at $\delta=0.95$, averaged over 25 random seeds.}
\centering
\label{tab:creg_wheel}
\begin{tabular}{lcc}
\toprule
Algorithm & $k$ & Mean $\pm$ SE \\
\midrule
NeuralLinear & N/A & $30266.0 \pm 6572.7$ \\
MAP (no exploration) & N/A & $43298.9 \pm 7146.9$ \\
Subnet Diagonal & $100$ & $47184.9 \pm 7275.3$ \\
Subnet Diagonal & $500$ & $33796.4 \pm 7464.6$ \\
Last $k$ & $100$ & $35340.1 \pm 7576.7$ \\
Last $k$ & $500$ & $30556.2 \pm 7942.9$ \\
Gradient-Laplace & $100$ & $38177.4 \pm 6904.8$ \\
\textbf{Gradient-Laplace} & $500$ & $8211.9 \pm 355.0$ \\
Greedy-Laplace & $100$ & $21568.7 \pm 5895.4$ \\
\textbf{Greedy-Laplace} & $500$ & $8616.7 \pm 574.5$ \\
\bottomrule
\end{tabular}
\end{table}

\subsection{Sensitivity to Sparsity ($\delta=0.9$)}

To assess how the comparison depends on the sparsity parameter, we repeat the experiment with $\delta=0.9$, which enlarges the high-reward region from $9.75\%$ to $19.0\%$ of the disk. All other settings are unchanged.
\begin{figure}[h]
    \centering
    \includegraphics[width=0.78\linewidth]{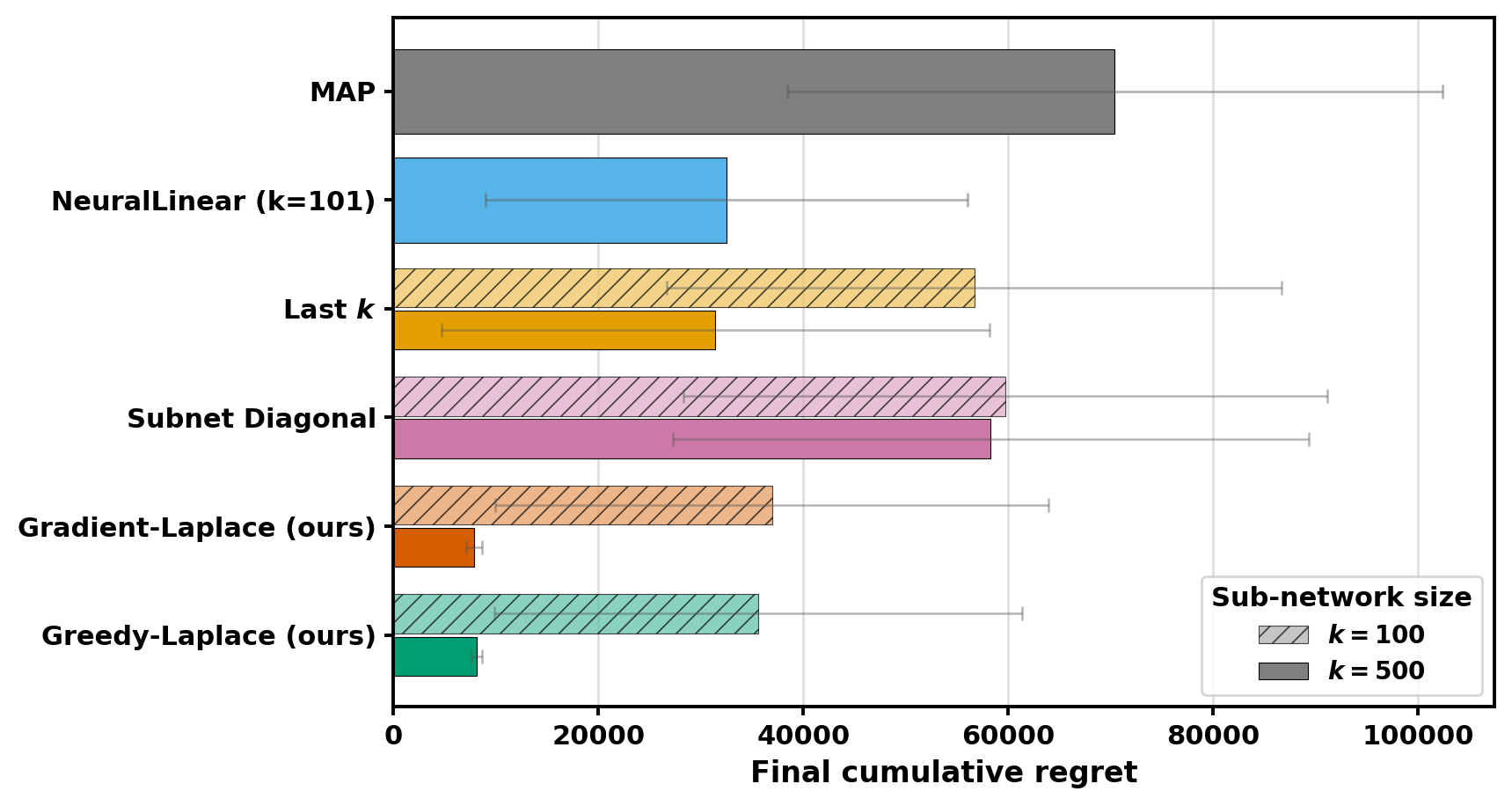}
    \caption{Final cumulative regret on the Wheel Bandit at $\delta=0.9$ over $T=16{,}000$ rounds. Bars show means over 25 random seeds, with error bars equal to $\pm 1.96 \times \mathrm{SE}$. Lower is better. For sub-network methods, paired bars correspond to $k=100$ and $k=500$; \emph{NeuralLinear} and MAP do not depend on $k$.}
    \label{fig:wheel_d09}
\end{figure}
Figure~\ref{fig:wheel_d09} reports final cumulative regret at $\delta=0.9$, and Table~\ref{tab:creg_wheel_d09} gives the corresponding numerical values.

\begin{table}[hbtp]
\caption{Final cumulative regret on the Wheel Bandit at $\delta=0.9$, averaged over 25 random seeds.}
\centering
\label{tab:creg_wheel_d09}
\begin{tabular}{lcc}
\toprule
Algorithm & $k$ & Mean $\pm$ SE \\
\midrule
NeuralLinear & N/A & $26745.9 \pm 13956.9$ \\
MAP (no exploration) & N/A & $50082.2 \pm 19103.3$ \\
Subnet Diagonal & $100$ & $60640.0 \pm 20071.5$ \\
Subnet Diagonal & $500$ & $48953.1 \pm 21940.2$ \\
Last $k$ & $100$ & $38473.7 \pm 16114.2$ \\
Last $k$ & $500$ & $32131.9 \pm 17366.6$ \\
Gradient-Laplace & $100$ & $20747.6 \pm 12895.4$ \\
\textbf{Gradient-Laplace} & $500$ & $7654.6 \pm 430.0$ \\
Greedy-Laplace & $100$ & $27527.0 \pm 13802.1$ \\
\textbf{ Greedy-Laplace} & $500$ & $7732.3 \pm 383.1$ \\
\bottomrule
\end{tabular}
\end{table}

The qualitative ordering at $\delta=0.9$ is consistent with the $\delta=0.95$ result in Section~\ref{subsec:wheel_bandit}: the proposed methods substantially outperform the existing sub-network baselines, and \emph{Subnet Diagonal} again incurs regret comparable to that of MAP, in line with Theorem~\ref{thm2}.

\end{document}